\documentclass[accepted]{uai2025} 
                        
\usepackage[american]{babel}

\usepackage{natbib} 
\usepackage{mathtools} 
\usepackage{booktabs} 
\usepackage{tikz} 

\usepackage{amsmath,amssymb,amsfonts}

\usepackage{graphicx}
\usepackage{textcomp}
\usepackage{xcolor}

\usepackage{amssymb}
\usepackage{amsmath}
\usepackage{amsthm}
\usepackage{booktabs}
\usepackage{enumitem}
\usepackage{graphicx}
\usepackage{color}

\usepackage{amsmath}
\usepackage{amsthm}
\usepackage{booktabs}
\usepackage{algorithm}
\usepackage[switch]{lineno}

\usepackage{algorithm}
\usepackage{algorithmic}
\usepackage{amsmath}
\usepackage{amssymb}
\usepackage{accents}
\usepackage{mathtools}
\usepackage{makecell}
\usepackage{multirow}
\usepackage{arydshln}
\usepackage{subcaption}

\usepackage{bm}

\newlength{\dhatheight}
\newcommand{\doublehat}[1]{%
    \settoheight{\dhatheight}{\ensuremath{\hat{#1}}}%
    \addtolength{\dhatheight}{-0.15ex}%
    \hat{\vphantom{\rule{1pt}{\dhatheight}}%
    \smash{\hat{#1}}}}
\newcommand{\summation}[2]{\sum\limits^{#1}_{#2}}

\usepackage{color,soul}
\usepackage{xcolor}
\definecolor{cerulean}{rgb}{0.0,0.48,0.65}
\definecolor{green}{rgb}{0.01, 0.75, 0.24}
\definecolor{Black}{RGB}{0.0, 0.0, 0.0}

\newcommand{\olive}[1]{\textcolor{Black}{#1}}
\newcommand{\br}[1]{\textcolor{Black}{#1}}

\newcommand{\teal}[1]{\textcolor{Black}{#1}}
\newcommand{\brown}[1]{\textcolor{Black}{#1}}
\newcommand{\shadow}[1]{}
\renewcommand{\arraystretch}{1.15}

\def\o{\olive}
\def\s{\shadow}

\def\t{\teal}
\def\br{\brown}

\title{Contaminated Multivariate Time-Series Anomaly Detection with Spatio-Temporal Graph Conditional Diffusion Models}

\author[1,2]{Thi Kieu Khanh Ho}
\author[1,2]{Narges Armanfard}

\affil[1]{%
   Department of Electrical and Computer Engineering, McGill University
}
\affil[2]{%
   Mila - Quebec AI Institute, Montreal, QC, Canada
}

\begin{document}
\maketitle

\begin{abstract}
Mainstream unsupervised anomaly detection algorithms often excel in academic datasets, yet their real-world performance is restricted due to the controlled experimental conditions involving clean training data. Addressing the challenge of training with noise, a prevalent issue in practical anomaly detection, is frequently overlooked.
In a pioneering endeavor, this study delves into the realm of label-level noise within sensory time-series anomaly detection (TSAD). 
This paper presents a novel and practical TSAD when the training data is contaminated with anomalies.  
The introduced approach, called TSAD-C, is devoid of access to abnormality labels during the training phase. TSAD-C encompasses three modules: a Decontaminator to rectify anomalies present during training and swiftly prepare the decontaminated data for subsequent modules; a Long-range Variable Dependency Modeling module to capture long-range intra-variable and inter-variable dependencies within the decontaminated data that is considered as a surrogate of the pure normal data; and an Anomaly Scoring module that leverages insights of the first two modules to detect all types of anomalies.
Our extensive experiments conducted on four reliable, diverse, \br{and challenging} datasets conclusively demonstrate that TSAD-C surpasses existing methods, thus establishing a new state-of-the-art in the TSAD field.
\end{abstract}

\section{Introduction} \label{sec:introduction}

Multivariate time-series data (MTS) is referred to as time-stamped data that consists of multiple variables, i.e., for each timestamp, there are multiple values associated with it. Time-series anomaly detection (TSAD) is the process of detecting unusual patterns or events within time-series data that deviate from the expected behavior \cite{schmidl2022anomaly}.  The unusual patterns can be found in many real-world applications such as instances of financial fraud, abrupt temperature spikes or unforeseen precipitation in weather data, security breaches, system malfunctions, and irregularities in brain activities. Many algorithms have been proposed to detect anomalies in time-series data \cite{su2019robust,audibert2020usad,deng2021graph,chen2022deep,yang2023dcdetector,ho2023self,chen2024imdiffusion,sun2024unraveling,fang2024temporal}. However, several critical challenges remain unresolved.
%

First, existing studies can be categorized into two groups, which are supervised methods utilizing both labeled normal and abnormal data in the training phase \cite{lai2023open}, and unsupervised methods assuming that only normal data is available during training \cite{yang2023dcdetector}. Given the demanding nature of accurately labeling anomalous patterns -- proving to be time-consuming, costly, and labor-intensive -- supervised approaches are deemed impractical. Consequently, there has been a recent surge in the development of unsupervised approaches. However, in many real-world applications, anomalies often sneak into normal data, which come from the data shift or human misjudgment \cite{jiang2022softpatch}. These unsupervised methods are sensitive to the seen anomalies due to their exhaustive strategy to model the normal training data, hence, they would misdetect similar anomaly samples in the test phase. Therefore, developing a method that can detect anomalies while being trained on contaminated data is necessary, yet there is no method that has aimed to tackle this challenge in the TSAD field \textbf{(i)}.

Second, it is important to capture both intra-variable (aka temporal) and inter-variable (aka spatial) dependencies in MTS \cite{ho2023graph}. However, existing studies are unable to effectively capture them. Regarding the intra-variable dependencies, they often preprocessed data by segmenting the signals into short time intervals \cite{lai2023open}, or applied conventional networks such as recurrent neural networks or transformers \cite{yang2023dcdetector}. Such learned dependencies imply that observations close to each other are expected to be similar. This is problematic as trends, seasonality and unpredictability are always present in MTS \cite{ho2023graph}. Thus, developing a method that can handle the \emph{long-range} intra-variable dependencies to aid in distinguishing normal and abnormal variations in MTS is crucial \textbf{(ii)}.

Regarding the inter-variable dependencies, very recently, graphs have brought the potential to model the relationships between variables (sensors) in MTS. By representing variables as nodes and their connections as edges, graphs provide an intuitive way to understand the underlying relationships between variables - a useful property in TSAD. For example, the changes in one variable can be used to predict the changes in another if they are correlated. However, modeling graphs to effectively capture this dependency type is challenging. Existing studies proposed to pre-define graphs based on prior knowledge, e.g., the known locations of sensors \cite{tang2021self,ho2023self,hojjati2023multivariate}. However, in many real-world applications, pre-defining the graph properties such as node locations and node features is not practical due to dynamic testing environments -- e.g., electroencephalogram sensor locations in comatose/epilepsy patients may vary depending on the brain damaged regions. Hence, instead of pre-defining graphs, dynamically learning the graph over time is highly desirable.

Lastly, while long-range intra-variable and inter-variable dependencies are both important for TSAD, designing an effective joint learning framework to capture them is yet challenging. Recent studies have shown that there are two groups of a combined model: time-and-graph and time-then-graph \cite{gao2022equivalence}. The time-and-graph approach first constructs graphs and then embeds a temporal network. On the other hand, the time-then-graph framework first projects time-series data to a temporal network, the extracted temporal features are then used to model graphs. It is shown that compared to the time-and-graph framework, the time-then-graph approach achieves a significant improvement in classification and regression tasks \cite{gao2022equivalence,tang2023modeling}. Yet, till now, no study has explored the time-then-graph framework for unsupervised TSAD \textbf{(iii)}.

Based on the above observations we propose a novel approach, called TSAD-C, that addresses \textbf{(i)}-\textbf{(iii)} challenges. The Related Work section is provided in \textbf{Appendix \ref{sec:related_works}}. The main contributions of this paper are described as follows:

\begin{itemize}
    \item We propose a novel \textit{fully} unsupervised approach, namely TSAD-C, trained on contaminated data in an end-to-end manner to detect all types of anomalies in MTS. To the best of our knowledge, this is the first study that uses contaminated data in the training phase for TSAD, addressing a much more challenging problem than the existing studies. 
    \item TSAD-C consists of three modules, namely Decontaminator, Long-range Variable Dependency Modeling, and Anomaly Scoring. The initial module aims to identify and eliminate abnormal patterns that are likely to be anomalies. This step results in decontaminated data, which is prepared swiftly for subsequent modules. The second module is a time-then-graph approach that is designed to model the long-range intra- and inter-variable dependencies within the decontaminated data. The last module computes anomaly scores to detect anomalies.
    \item The novel Decontaminator employs masking strategies and a structure state space (S4)-based conditional diffusion model, while the second module integrates Intra-variable Modeling, and Inter-variable Modeling components. The Anomaly Scoring module leverages insights of the first two modules.
    \item Extensive experiments on four reliable, diverse, \br{and challenging} datasets demonstrate that our method outperforms existing studies, thus establishing a new state-of-the-art in the TSAD field.
\end{itemize}

\section{Proposed Method} \label{sec:method}

A dataset is defined as $X = (\mathbf{x}_{(1)}, \mathbf{x}_{(2)}, \ldots, \mathbf{x}_{(N)})$, where $\mathbf{x}_{(i)}  = (x_{(i)}^1,x_{(i)}^2,\ldots, x_{(i)}^K)$ is the $i$th observation in the time series of $N$ observations, $\mathbf{x}_{(i)} \in \mathbb{R}^{K \times L}$, $K$ and $L$ denote the number of variables (sensors) and the length of the $i$th observation, respectively. \s{An observation can be conceptualized as $L$ samples collected from $K$ sensors over the $i$th time interval. }Our task is to detect anomalous observations from all types in the test data $X_{\text{test}}$ by training the model with contaminated data $X_{\text{train}}$. No information about anomalies that contaminate normal data is provided during training, such as their labels or their positions within the time series. A validation set $X_{\text{valid}}$ is used for early stopping and finding the decision threshold. 
The block diagram of the proposed TSAD-C method, depicting the three modules, is shown in Figure \ref{fig:model}. Details of each module are presented below. The pseudocodes are provided in \br{\textbf{Appendix \ref{sec:pseudocodes}}}. The source code is also available in the provided zip file.

\begin{figure*}[htbp]
\centering
\includegraphics[width=0.9\linewidth]{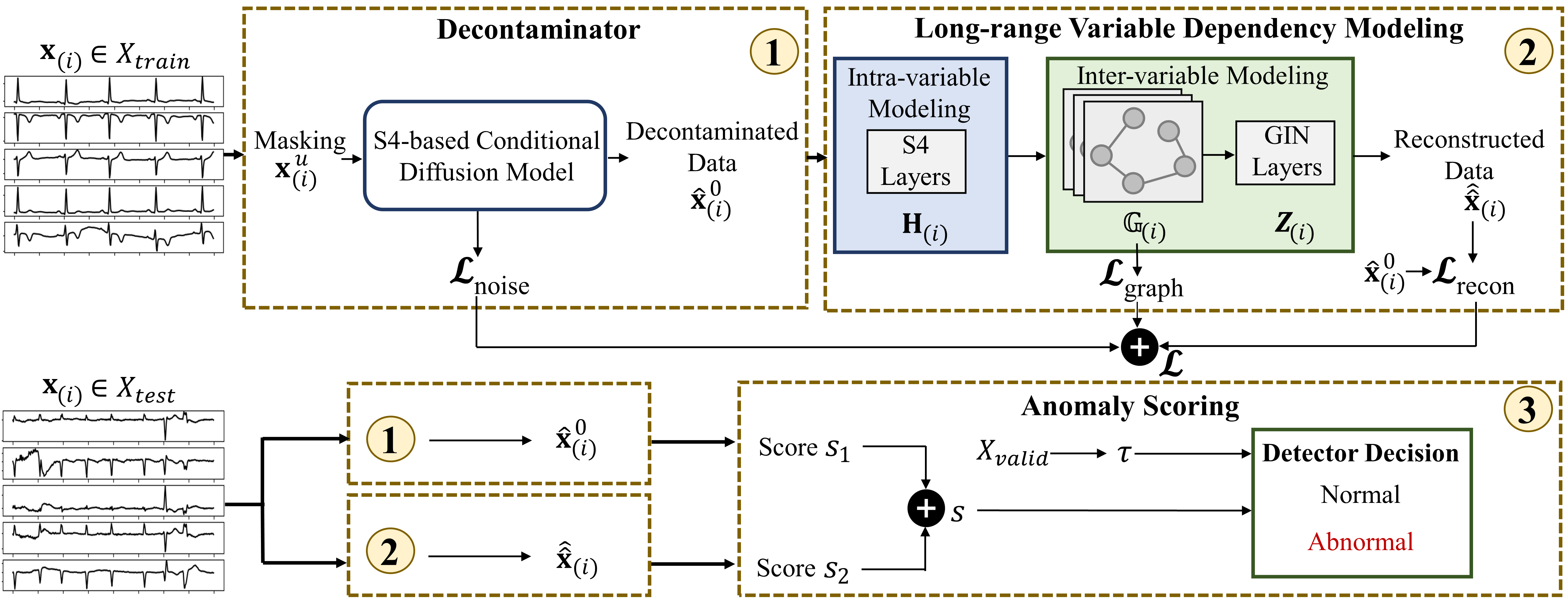}
\caption{The overall framework of TSAD-C consists of three modules: the Decontaminator integrates masking strategies and an S4-based diffusion model, the Long-Range Variable Dependency Modeling module incorporates Intra- and Inter-variable Modeling components; and the Anomaly Scoring module leverages insights from the preceding modules to detect anomalies.}
\label{fig:model}
\end{figure*}

\subsection{Decontaminator} \label{subsec:decontaminator}

This module incorporates masking strategies and an S4-based conditional diffusion model. As no information about anomalies is provided during training, we propose masking strategies to decontaminate the input data. The diffusion model is then deployed to rectify anomalies, with S4 - a noise estimator included to ensure that long-range intra-variable dependencies are effectively captured. Notably, we introduce a pioneering concept in the diffusion field, i.e., minimizing the noise error on masked portions for a simpler and more streamlined training process. The decontaminated data is then obtained by a \emph{single} step during the reverse process, which is a fast data preparation for subsequent modules – a significant advantage for practical applications.

\textbf{Masking Strategies.} 
Following the practical scenarios, we assume that normal samples significantly outnumber anomalies. When masking a portion of $X_{\text{train}}$, both normal and abnormal patterns might be removed. As  normal data predominates in $X_{\text{train}}$, omitting some normal patterns is not likely to yield detrimental consequences as the substantial amount of remaining normal data can compensate for masked portions. Conversely, masking can help reducing the proportion of anomalies. This benefits the downstream module as it facilitates the learning process of variable dependencies that characterize the underlying behavior of normality.

We define a mask as $\mathbf{v} \in \{0,1\}$, $\mathbf{v} \in \mathbb{R}^{K \times L}$, where zeros and ones denote the values to be masked and the values to be kept, respectively. Hence, the $i$th masked observation is $\mathbf{x}_{(i)}^u = \mathbf{x}_{(i)} \odot \mathbf{v}$, where $\mathbf{x}_{(i)}^u \in \mathbb{R}^{K \times L}$ and $\odot$ denotes point-wise multiplication. We perform three masking scenarios, namely random masking (RandM), random block masking (RandBM) and blackout masking (BoM) \cite{alcaraz2022diffusion}. We control the masking ratio by the hyperparameter $r$, which specifies the number of timestamps to be masked. RandM randomly samples $r$ to be masked across variables. In RandBM, there might be no time overlap between the masked windows across variables, whereas in BoM, the same time window is masked across all variables. Note that each masked window has the size of $r$.

\textbf{S4-based Conditional Diffusion Model.} This is developed based on a diffusion model\s{, a class of generative models inspired by non-equilibrium thermodynamics} \cite{croitoru2023diffusion} that includes the diffusion and reserve processes. In this paper, the diffusion process\s{, defined by a Markov chain,} incrementally adds Gaussian noise to the initial stage of $\mathbf{x}^u_{(i)}$, called $\mathbf{x}_{(i)}^0$, over $T$ diffusion steps:
\begin{equation}
    p(\mathbf{x}_{(i)}^1, \ldots, \mathbf{x}_{(i)}^T | \mathbf{x}_{(i)}^0) = \prod_{t=1}^T p(\mathbf{x}_{(i)}^{t}|\mathbf{x}_{(i)}^{t-1}),
\end{equation}
where $p(\mathbf{x}_{(i)}^t|\mathbf{x}_{(i)}^{t-1}) \coloneqq \mathcal{N}(\mathbf{x}_{(i)}^t; \mu^t_{(i)}, \sigma^t_{(i)})$.  This indicates that $\mathbf{x}_{(i)}^t$ is sampled from a normal distribution with mean $\mu^t_{(i)} = \sqrt{1-\beta_t}\mathbf{x}_{(i)}^{t-1}$ and variance $\sigma^t_{(i)}=\beta_t \mathbf{I}$. $\mathbf{I}$ is the identity matrix, $\beta_t \in (0,1)$ is a variance scheduler that controls the quantity of noise added at the $t$th diffusion step. In our implementation, we increase $\beta_t$ linearly from $10^{-4}$ to $0.02$. By setting $\alpha_t = 1 - \beta_t$, $\Bar{\alpha}_t = \prod_{j=1}^t \alpha_j$, the diffusion process allows to immediately transform $\mathbf{x}_{(i)}^0$ to a noisy $\mathbf{x}_{(i)}^t$ according to $\beta_t$ in a closed form as $\mathbf{x}_{(i)}^t = \sqrt{\Bar{\alpha}_t}\mathbf{x}_{(i)}^0 + \sqrt{1-\Bar{\alpha}_t}\epsilon_t$ where the noise $\epsilon_t \sim \mathcal{N}(0,\mathbf{I})$, $\epsilon_t \in \mathbb{R}^{K \times L}$. We add noise to both masked and non-masked portions of $\mathbf{x}_{(i)}^0$. As the diffusion step increases, $\mathbf{x}_{(i)}^0$ gradually loses its distinguishable features and approaches a Gaussian distribution; hence, both anomalous and normal patterns appear indistinguishable.

The reverse process \s{defined by a Markov chain} is parameterized by $\theta$ as:
\begin{equation} \label{eq:p_sampling}
    q_{\theta}(\mathbf{x}_{(i)}^0, \ldots, \mathbf{x}_{(i)}^{T-1}| \mathbf{x}_{(i)}^T) = \prod_{t=1}^T q_{\theta}(\mathbf{x}_{(i)}^{t-1}|\mathbf{x}_{(i)}^t),
\end{equation}
where each 
$q_{\theta}(\mathbf{x}_{(i)}^{t-1}|\mathbf{x}_{(i)}^t) \coloneqq$ $\mathcal{N}(\mathbf{x}_{(i)}^{t-1}; \mu_{\theta}(\mathbf{x}_{(i)}^t,t,c),$  $\sigma_{\theta}(\mathbf{x}_{(i)}^t,t,c)^2\mathbf{I})$. $\mu_{\theta}$ and $\sigma_{\theta}$ are parameterized as:
\begin{equation} \label{eq:sampling}
    \begin{array}{cc}
    \mu_{\theta}(\mathbf{x}_{(i)}^t,t,c) = \frac{1}{\sqrt{\alpha_t}} \Big(\mathbf{x}_{(i)}^t-\frac{\beta_t}{\sqrt{1-\Bar{\alpha}_t}} \epsilon_{\theta}(\mathbf{x}_{(i)}^t,t,c)  \Big), \\
    \sigma_{\theta}(\mathbf{x}_{(i)}^t,t,c) = \sqrt{\Bar{\beta}_t},
    \end{array}
\end{equation}
where $\Bar{\beta}_t = \frac{1-\Bar{\alpha}_{t-1}}{1-\Bar{\alpha}_t}\beta_t$ and $\Bar{\beta}_1 = \beta_1$. $\epsilon_{\theta}$ is a noise estimator, which takes $\mathbf{x}_{(i)}^t$, the diffusion step $t$ and a conditional factor $c$ as the inputs and aims to predict the noise from $\mathbf{x}_{(i)}^t$. \s{Compared to the unconditional diffusion models, conditional cases typically achieve a better performance as the reverse process is conditioned on additional information \cite{alcaraz2022diffusion}. In our scenario, }$c$ is a concatenation of non-masked parts in $\mathbf{x}_{(i)}^u$ and the positional information of masked parts provided by $\mathbf{v}$. This extra information facilitates our reverse process to distinguish the zero portions of non-masked and masked parts.

\begin{figure}[t]
\centering
\includegraphics[width=1.01\linewidth]{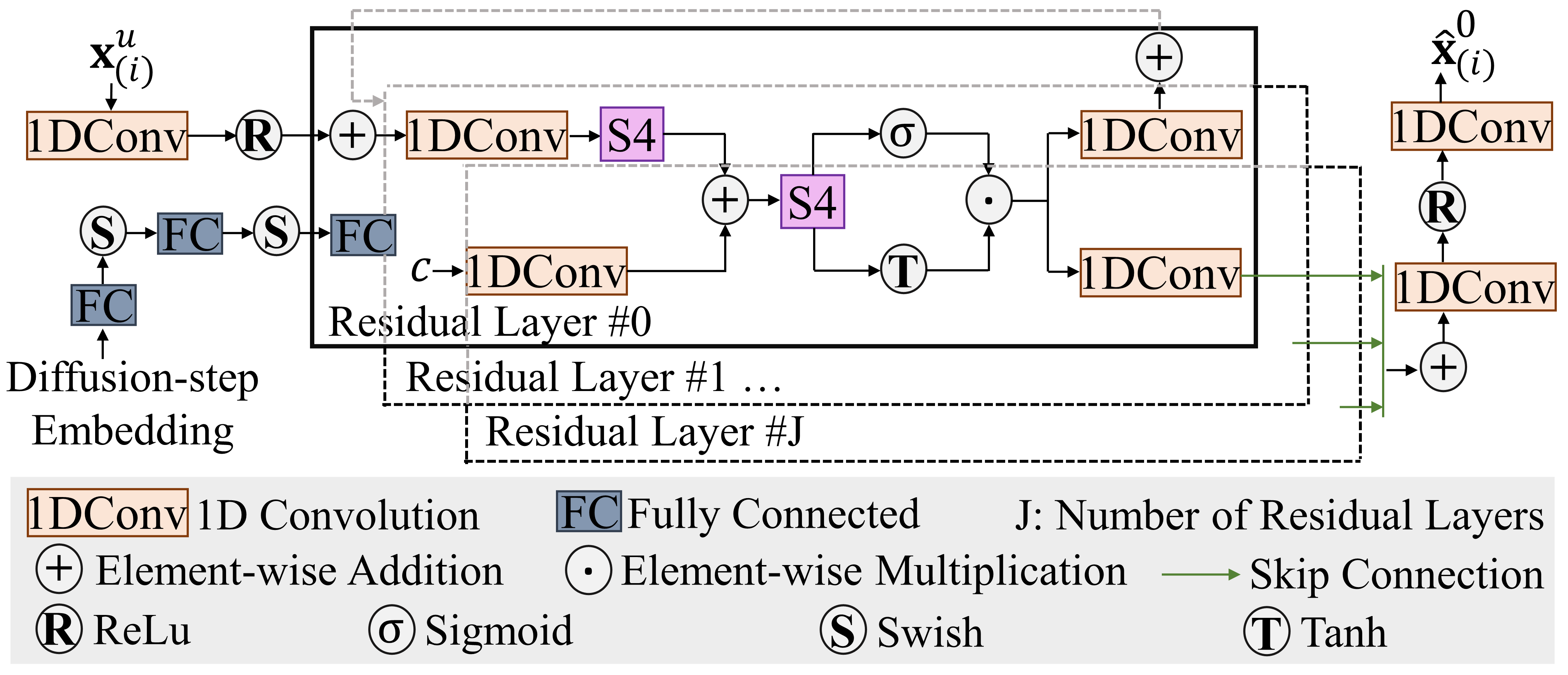}
\caption{\o{The architecture of the Decontaminator includes two S4 layers in every residual block to ensure that long-range intra-variable dependencies are effectively captured.}}
\label{fig:decontminator}
\end{figure}

Note that $\epsilon_{\theta}$ plays a key role in our reverse process. \s{U-Net-like architectures, based on PixelCNN and ResNet, are often utilized to build $\epsilon_{\theta}$ in the image domain \cite{zhang2023diffusionad}. However, in the time-series domain,}Since capturing long-range intra-variable dependencies is crucial, we propose to build $\epsilon_{\theta}$ based on S4 \cite{gu2021efficiently} - a recent deep sequence model with the concept of a state space model (SSM). A continuous-time SSM maps $\mathbf{x}_{(i)}^t$ to a high dimensional state $h_{(i)}^t$ before projecting it to the output $\mathbf{y}_{(i)}^t$. This transition  can be defined  as:
\begin{equation}
    \Tilde{h}^t_{(i)} = A h_{(i)}^t + B\mathbf{x}_{(i)}^t \text{  and  } \mathbf{y}_{(i)}^t = C h_{(i)}^t + D\mathbf{x}_{(i)}^t,
\end{equation}
where $A,B,C,D$ are transition matrices learned by gradient descent. However, S4 shows that a discrete-time SSM can be represented as a convolution operation by:
\begin{equation} \label{eq:discretized_eq}
    \overline{O}\coloneqq (\overline{CB}, \overline{CAB}, \ldots, \overline{CA}^{L-1}\overline{B}) \text{  ,  } \mathbf{y}_{(i)} = \overline{O} \ast \mathbf{x}_{(i)}, 
\end{equation}
where $\overline{A},\overline{B},\overline{C}$ are the discretized matrices, $\overline{CA}^{L-1}$ denotes the multiplication of discretized matrices at $L-1$, and $\overline{O}$ is a SSM convolution kernel. $D$ is omitted in Equation \eqref{eq:discretized_eq} as $D\mathbf{x}_{(i)}^t$ can be viewed as a skip connection. S4 parameterizes $A$ as a diagonal plus low rank matrix that enables fast computation of $\overline{O}$. It also includes the HiPPO matrices \cite{gu2020hippo} capable of capturing long-term intra-variable dependencies.
We employ two S4 layers, one after the addition of the embeddings related to $\mathbf{x}_{(i)}^u$ and another layer after including $c$ in the residual blocks, shown in Figure \ref{fig:decontminator}.

To have a simpler and more streamlined reverse process during the training of $\epsilon_\theta$, we suggest minimizing the noise error on the masked parts shown in Equation \eqref{eq: loss_noise}. Note that the masked parts only consist of noise without any actual data patterns, unlike the non-masked parts containing both noise and the actual data patterns. Hence, deriving noise estimated from the masked data is a more straightforward task and can be accomplished using less intricate networks. This approach also speeds up data preparation for subsequent modules. We provide \br{\textbf{Section \ref{subsec:decontaminator_study}}} that compares two optimization strategies: minimizing $\mathcal{L}_\text{noise}$ on the masked portions versus minimizing $\mathcal{L}_\text{noise}$ on the entire observations as done by prior diffusion studies \cite{alcaraz2022diffusion,chen2024imdiffusion}. It shows that the former approach yields superior performance, simplicity and applicability. 
\begin{equation} \label{eq: loss_noise}
    \mathcal{L}_{\text{noise}} = \|\epsilon_t \odot (1-\mathbf{v})- \hat{\epsilon}_t \odot (1-\mathbf{v}) \|^2,
\end{equation}
where $\hat{\epsilon}_t$ is the predicted noise obtained from $\epsilon_{\theta}(\mathbf{x}_{(i)}^t,t,c)$. 

We then obtain the decontaminated data in the training as:
\begin{equation}
    \hat{\mathbf{x}}_{(i)}^0 = \frac{1}{\sqrt{\Bar{\alpha}_T}} \Big(\mathbf{x}_{(i)}^T - \sqrt{1-\Bar{\alpha}_T} \hat{\epsilon}_T\Big).
\end{equation}
Note that approximating  $\hat{\mathbf{x}}_{(i)}^0$ by a \emph{single} step (immediately at the $T$th step) enables a faster reverse speed as $\hat{\mathbf{x}}_{(i)}^0$ serves as the input for the second module during training. Meanwhile, during testing, we perform a complete sampling step from $T$ to 1 based on Equations \eqref{eq:p_sampling} and \eqref{eq:sampling} to obtain $\hat{\mathbf{x}}_{(i)}^0$.

\textbf{Theoretical Analysis of Decontamination Effectiveness.}
Reducing the fraction of anomalies (\emph{``impurity''}) in the training data is crucial for accurate anomaly detection. Training on less contaminated data helps the model learn normal patterns more effectively, thereby improving its ability to detect genuine anomalies.

We define $N$ as the total number of training samples, $N_a$ as the number of truly anomalous samples, $\eta_a = \frac{N_a}{N}$ as the \emph{true} anomaly ratio (with $0 < \eta_a < 0.5$), $\eta_e \in (0,1)$ as the \emph{estimated} anomaly ratio chosen by the user, and $ip_1 = \eta_a$ as the initial impurity level.

\textit{Impurity Reduction via Decontaminator Masking.}
During training, the Decontaminator masks a fraction $\eta_e$ of all samples \emph{uniformly at random}. Hence, on average, $\eta_e N_a$ anomalous samples are also masked. Since $0 < \eta_a < 0.5$, the new impurity level after decontamination is:
\[
    ip_2 
    = \frac{N_a - \eta_e N_a}{N} 
    = \eta_a \,(1 - \eta_e).
\]
Since $0 < \eta_e < 1$, we have
\[
    ip_2 
    = \eta_a \,(1 - \eta_e) 
    < \eta_a 
    = ip_1.
\]
Thus, the impurity always \emph{strictly decreases} compared to the original contamination level.

\textit{User-Controlled Impurity Reduction.}
Users can \emph{control} how much to reduce the impurity by selecting $\eta_e$. Suppose one wishes to reduce the original impurity $ip_1=\eta_a$ to a fraction $\alpha \in (0,1)$:
\[
    ip_2 
    = \alpha \, ip_1 
    = \alpha\,\eta_a.
\]
Solving $\eta_a(1-\eta_e) = \alpha\,\eta_a$ yields:
\[
    \eta_e = 1 - \alpha.
\]
For instance, if $\alpha = 0.5$, then one sets $\eta_e=0.5$ to remove $50\%$ of the initial impurity.

\textit{Robustness to Underestimation and Overestimation of $\eta_a$.}
In practice, $\eta_a$ is unknown. Nevertheless, the Decontaminator remains robust whether $\eta_e$ is an underestimate or overestimate.

Regarding underestimation ($\eta_e < \eta_a$),
we still have $ip_2 = \eta_a(1-\eta_e) < \eta_a = ip_1$. More specifically,
\[
    ip_1 - ip_1^2 
    < ip_2 
    < ip_1.
\]
Hence, even if one underestimates the contamination level, the Decontaminator still \emph{strictly} reduces impurity.

Regarding overestimation ($\eta_e \ge \eta_a$), we have 
\[
    ip_2 
    = \eta_a(1 - \eta_e) 
    \le \eta_a(1 - \eta_a) 
    = ip_1 - ip_1^2 \le ip_1.
\]
Since $\eta_a < 0.5$, the maximum of $ip_1 - ip_1^2$ over $\eta_a \in (0,0.5)$ is $0.25$. Thus, $ip_2 \le 0.25$. Even when a large fraction of samples is masked (e.g., $\eta_e=0.7$), as long as sufficient normal data remains, the model can still learn normal structure effectively.

\textit{Conclusion.}
Regardless of how $\eta_e$ relates to $\eta_a$, the Decontaminator \emph{strictly reduces} the contamination level in the training set ($ip_2 < ip_1$). This guarantees a cleaner dataset, where the model’s representations are primarily driven by normal patterns. As a result, reconstruction-based anomaly detection becomes more robust, even when $\eta_a$ is inaccurately estimated.

\subsection{Long-range Variable Dependency Modeling} \label{subsec:modeling}

This module builds a time-then-graph framework, motivated in \br{\textbf{Section 1}-Paragraph 5}, by incorporating two components: Intra-variable Modeling and Inter-variable Modeling.

\textbf{Intra-variable Modeling.} We propose to leverage multiple back-to-back S4 layers to capture long-range intra-variable dependencies. Specifically, we use $\hat{\mathbf{x}}_{(i)}^0 \in \mathbb{R}^{K \times L}$ as the input and project it onto an embedding space, called $\mathbf{H}_{(i)} \in \mathbb{R}^{K \times \Gamma \times U}$, where  $\Gamma$ and $U$ are hyperparameters defining the S4 embedding dimension. 
To maintain a sense of the number of timestamps present in $\mathbf{x}_{(i)}$, we set $\Gamma$ to $L$, corresponding to the input length.
This allows us to model long-term intra-variable dependencies within each variable. $\mathbf{H}_{(i)}$ is then used in the graph learning phase to model inter-variable dependencies. 
Prior studies have shown the superior performance of graph learning when using the temporal embedding rather than the original data \cite{tang2023modeling}.

\textbf{Inter-variable Modeling.} We represent $\mathbf{H}_{(i)}$ as a set of graphs $\mathbb{G}_{(i)} = \{\mathcal{G}_{(i)}^m\}_{m=1}^d$, where $d = \frac{\Gamma}{g}$ and $g$ is the pre-defined length of the short and non-overlapping time windows within the $i$th observation.
Since each observation can encompass thousands of timestamps, constructing a graph for every time step becomes inefficient and computationally demanding. Hence, we create a graph over a defined time window, which aids in information aggregation. This strategy not only leads to a graph with reduced noise but also facilitates faster computations  \cite{gao2022equivalence}.

We define $\mathcal{G}_{(i)}^m = \{\textbf{E}_{(i)}^m, \mathcal{A}_{(i)}^m\}$. $\textbf{E}_{(i)}^m \in \mathbb{R}^{K \times U}$ denotes the embedding derived by averaging the elements of $\mathbf{H}_{(i)}$ along its second dimension. $\mathcal{A}_{(i)}^m \in \mathbb{R}^{K \times K}$ is the adjacency matrix. Each row and column in $\mathcal{A}_{(i)}^m$ correspond to a node (variable). The non-zero value indicates that there exists an edge connecting the two nodes. We then employ a self-attention paradigm \cite{tang2023modeling} in which attention weights are assigned to the edges' weights, represented as: $    \mathbf{Q} = \textbf{E}_{(i)}^m \mathbf{W}^{\mathbf{Q}} \text{,  } \mathbf{R} = \textbf{E}_{(i)}^m \mathbf{W}^{\mathbf{R}}, 
     \mathcal{A}_{(i)}^m = \text{softmax} (\frac{\mathbf{Q} \mathbf{R}^{\top}}{\sqrt{D}}),$
%
where $\mathbf{W}^{\mathbf{Q}}, \mathbf{W}^{\mathbf{R}} \in \mathbb{R}^{U \times U}$ are the learnable weights that project $\textbf{E}_{(i)}^m$ to the query $\mathbf{Q}$ and the key $\mathbf{R}$, respectively.

To help guiding the graph learning process, \s{In addition to the learnable parameters for creating $\mathcal{A}_{(i)}^m$,} we also include a pre-defined adjacency matrix, called $\mathcal{A'}_{(i)}^{m}$, based $\delta$-nearest neighbors\s{ to help guiding the graph learning process}. Its edge values are computed by the cosine similarity between the nodes' embeddings in $\textbf{E}_{(i)}^m$. We keep the top $\delta$ edges that have the highest values for each node to avoid overly connected graphs. In our experiments, $\delta=3$.
Hence, the final $\mathcal{A}_{(i)}^m  = \zeta \mathcal{A'}_{(i)}^{m} + (1-\zeta)\mathcal{A}_{(i)}^m$, where the hyperparameter $\zeta \in [0,1)$ balances the two components.

It is important to regularize the graph to ensure desired graph properties such as smoothness (the features should change smoothly between neighboring nodes), sparsity (avoiding an overly connected graph) and connectivity (avoiding a disconnected graph) \cite{zhu2021survey}. Hence, we include three constraints in the regularization loss as: 
\begin{equation} \label{eq: loss_graph}
    \begin{aligned}
        \mathcal{L}_{\text{graph}} = \frac{1}{d} \sum_{m=1}^d \xi_1 \mathcal{L}_{\text{smooth}} (\textbf{E}_{(i)}^m, \mathcal{A}_{(i)}^m) + \\
        \xi_2 \mathcal{L}_{\text{sparse}}(\mathcal{A}_{(i)}^m)  + \xi_3 \mathcal{L}_{\text{connect}}(\mathcal{A}_{(i)}^m),
    \end{aligned}
\end{equation}
where $\mathcal{L}_{\text{smooth}} = \frac{1}{K^2} \text{tr} (\textbf{E}_{(i)}^{m\top} M_{\text{Lap}} \textbf{E}_{(i)}^{m})$, $M_{\text{Lap}} = M_{\text{degree}} - \mathcal{A}_{(i)}^m$ is the Laplacian matrix, $M_{\text{degree}}$ is the degree matrix of $\mathcal{A}_{(i)}^m$, and $\text{tr}(\cdot)$ denotes the trace. $\mathcal{L}_{\text{sparse}} = \frac{1}{K^2} \|\mathcal{A}_{(i)}^m\|_F^2$ and $\|\cdot\|_F$ is the Frobenius norm. $\mathcal{L}_{\text{connect}} = - \frac{1}{K} \textbf{1}^{\top} \log (\mathcal{A}_{(i)}^m \cdot \textbf{1})$, and $\textbf{1} \in \mathbb{R}^{K \times 1}$ is a matrix of ones. $\xi_1, \xi_2$ and $\xi_3$ are hyperparameters defined to balance the terms in $\mathcal{L}_{\text{graph}}$.

We then leverage a graph isomorphism network (GIN), which shows a strong representational power \cite{xu2018powerful} to capture inter-variable dependencies between nodes in $\mathcal{G}^m_{(i)}$. The embedding of nodes in $\mathcal{G}_{(i)}^m$ is represented as $\mathbf{z}_{(i)}^{m} \coloneqq \text{GIN} (\textbf{E}_{(i)}^m, \mathcal{A}_{(i)}^m)$, $\mathbf{z}_{(i)}^{m} \in \mathbb{R}^{K \times g \times U}$. We concatenate the node embeddings of all graphs within $\mathbb{G}_{(i)}$ as $\mathbf{Z}_{(i)} = \text{concat} (\mathbf{z}_{(i)}^1, \ldots, \mathbf{z}_{(i)}^d)$. Finally, a linear layer is added to obtain reconstructed data $\doublehat{\mathbf{x}}_{(i)} \coloneqq \text{Linear} (\mathbf{Z}_{(i)})$, $\doublehat{\mathbf{x}}_{(i)} \in \mathbb{R}^{K \times L}$. Thus, the reconstruction loss is denoted as: 
\begin{equation} \label{eq: loss_rec}
     \mathcal{L}_{\text{recon}} = \|\hat{\mathbf{x}}_{(i)}^0 - \doublehat{\mathbf{x}}_{(i)}\|^2. 
\end{equation}

The final loss in the training phase is defined as \s{the total loss of all three losses from Equations \eqref{eq: loss_noise}, \eqref{eq: loss_graph} and \eqref{eq: loss_rec}}:
\begin{equation}
    \mathcal{L} = \mathcal{L}_{\text{noise}} + \mathcal{L}_{\text{graph}} + \mathcal{L}_{\text{recon}}.
\end{equation}

\subsection{Anomaly Scoring} \label{subsec:scoring}

In the test phase, we compute an anomaly score based the root mean square error (RMSE) for each $\mathbf{x}_{(i)}$.  Specifically, we project $\mathbf{x}_{(i)}$ to the Decontaminator where masking strategies and the complete sampling step from $T$ to 1 are applied to obtain $\hat{\mathbf{x}}_{(i)}^0$. If $\mathbf{x}_{(i)}$ is an anomaly, the masked portions are expected to be inaccurately sampled.  
Additionally, instead of using $\hat{\mathbf{x}}_{(i)}^0$ as the input for the second module as done during training, $\mathbf{x}_{(i)}$ is directly used to obtain $\doublehat{\mathbf{x}}_{(i)}$ in the test phase. The assumption is that if $\mathbf{x}_{(i)}$ is an anomaly, the second module with the goal of achieving the flawless reconstruction of normal patterns would be unable to reconstruct it.  The final score $s$ for each $\mathbf{x}_{(i)}$ is computed as:
\begin{equation}
    \begin{array}{l}
        \small{s_1 = \Big(\frac{1}{L} \summation{L}{l=1} \summation{K}{k=1} \Big((\hat{\mathbf{x}}_{(i)}^0 - \mathbf{x}_{(i)}) \odot (1-\mathbf{v})\Big)^2\Big)^{0.5},} \\
        \small{s_2 = \Big(\frac{1}{L} \summation{L}{l=1} \summation{K}{k=1} \Big(\doublehat{\mathbf{x}}_{(i)} - \mathbf{x}_{(i)}\Big)^2\Big)^{0.5},} \\
        s = \lambda_1 s_1 + \lambda_2 s_2,
    \end{array}
\end{equation}
where $s_1$ and $s_2$ are, respectively, the RMSE scores obtained from the first and second modules. $\lambda_1$ and $\lambda_2$ are hyperparameters defined to ensure that scores from both modules fall within a similar numeric range. For a fair comparison, they are fixed across all experiments and datasets.

We conduct a decision threshold search on the \emph{unlabeled} $X_{\text{valid}}$, deviating from the TSAD practice where the threshold is selected on a \emph{labeled} set \cite{carmona2022neural,chen2022deep}. This approach is impractical for unsupervised applications where labeled data is unavailable. This may also lead to overfitting when labeled data is not sufficient to represent the distribution of anomalies, a common challenge in TSAD. In our case, we determine the threshold $\tau$ by a quantile approach \cite{kuan2017quantile} applied to anomaly scores obtained from the unlabeled $X_{\text{valid}}$. Specifically, we select a quantile based on a rough estimation of the percentage of normal data in $X_{\text{valid}}$, as provided by the dataset provider. For instance, if about 20\% of the data is contaminated, we set the quantile to 80\%. In the test phase, observations with $s$ above $\tau$ are detected as anomalies. Note that there is no overlap between $X_{\text{train}}$, $X_{\text{valid}}$, and $X_{\text{test}}$. \s{Overall, our threshold finding approach eliminates the need for labeled data, while mitigating the risk of overfitting by relying on distributional characteristics rather than specific samples.}

\section{Experiments}

\subsection{Experimental Settings}

This section introduces the datasets, baselines and evaluation metrics in our study. Implementation details are provided in \textbf{Appendix \ref{sec:implementation_details}}. Due to the diversity in dataset characteristics and dimensions, we present dedicated discussions on TSAD-C's computational cost and scalability in \textbf{Appendix \ref{sec:scalability}}.

\renewcommand{\arraystretch}{1.4}
\begin{table}[h]
\centering
\caption{The numbers of observations are shown in $X_{\text{train}}$, $X_{\text{valid}}$ and $X_{\text{test}}$ of each dataset, with the anomaly ratio $\eta$ in parentheses. $f_s$ is the sampling rate (in Hertz), $K$ is the number of variables, and $L$ is the length of an observation.}
\scalebox{0.79}{
\begin{tabular}{>{\centering\arraybackslash}p{12mm} | >{\centering\arraybackslash}p{22mm} | >{\centering\arraybackslash}p{22mm} | >{\centering\arraybackslash}p{22mm} | >{\centering\arraybackslash}p{8mm} | >{\centering\arraybackslash}p{6mm} | >{\centering\arraybackslash}p{10mm}}
     \hline
     \textbf{Dataset} & $X_{\text{train}}$ ($\eta$) & $X_{\text{valid}}$ ($\eta$) & $X_{\text{test}}$ ($\eta$) & $f_s$ & $K$ & $L$ \\ 
     \hline 
     SMD & 1624 (10.3\%) & 276 (7.97\%) & 416 (33.8\%) & $\frac{1}{60}$ & 38 & 600  \\ 
     \hline
     ICBEB & 910 (20.0\%) & 82 (20.7\%) & 222  (59.9\%) & 100 & 12 & 6,000  \\ 
     \hline
     DODH & 2515 (19.8\%) & 320 (21.8\%) & 310 (51.6\%) & 250 & 16 & 7,500  \\ 
     \hline
     TUSZ & 5275 (17.0\%) & 1055 (20.0\%) & 1581 (40.0\%) & 200 & 19 & 12,000  \\ 
     \hline
\end{tabular}}
\label{tab:dataset}
\end{table}

\renewcommand{\arraystretch}{1.22}
\setlength{\tabcolsep}{7pt}
\begin{table*}[t]
    \centering
    \caption{Comparison of existing methods and TSAD-C. The best and second-best scores are in bold and underlined.}
    \scalebox{0.76}{
    \begin{tabular}{cc|ccccccccccccc}
        \hline 
        \multirow{2}{*}{\rotatebox[origin=c]{90}{\textbf{Group}}} & \multirow{2}{*}{\textbf{Method}} & \multicolumn{3}{c}{\textbf{SMD}} & \multicolumn{3}{c}{\textbf{ICBEB}} & \multicolumn{3}{c}{\textbf{DODH}} & \multicolumn{3}{c}{\textbf{TUSZ}} & \textbf{Average} \\ \cmidrule(lr){3-5} \cmidrule(lr){6-8} \cmidrule(lr){9-11} \cmidrule(lr){12-14} \cmidrule(lr){15-15}
         & & F1 & Rec & APR & F1 & Rec & APR & F1 & Rec & APR & F1 & Rec & APR & F1 \\ \hline
         \multirow{4}{*}{\rotatebox[origin=c]{90}{\textbf{Intra-}}}  & USAD & 0.261 & 0.227 & 0.398 & 0.579 & 0.485 & 0.705 & 0.355 & 0.419 & 0.514 & 0.450 & 0.393 & 0.581 & 0.411 \\ 
         & LSTM-AE & 0.332 & 0.411 & 0.445 & 0.609 & 0.651 & 0.752 & 0.534 & 0.706 & 0.643 & 0.471 & 0.424 & 0.592 & 0.486 \\ 
         & S4-AE & 0.313 & 0.305 & 0.432 & \underline{0.664} & \underline{0.735} & 0.749 & \underline{0.625} & \underline{0.821} & \underline{0.713} & 0.527 & \underline{0.576} & 0.615 & 0.532 \\ 
         & DCdetector & 0.318 & 0.276 & 0.448 & 0.626 & 0.598 & 0.718 & 0.442 & 0.550 & 0.576 & 0.485 & 0.447 & 0.600 & 0.468 \\ \hline
         \multirow{3}{*}{\rotatebox[origin=c]{90}{\textbf{Inter-}}} & 
         GAE & 0.241 & 0.191 & 0.395 & 0.575 & 0.454 & 0.746 & 0.480 & 0.613 & 0.604 & 0.508 & 0.476 & 0.625 & 0.451 \\ 
         & GDN & 0.301 & 0.283 & 0.423 & 0.586 & 0.515 & 0.741 & 0.524 & 0.687 & 0.636 & 0.456 & 0.398 & 0.585 & 0.467 \\ 
         & EEG-CGS & 0.295 & 0.291 & 0.415 & 0.561 & 0.470 & 0.740 & 0.502 & 0.650 & 0.620  & 0.516 & 0.490 & 0.619 & 0.469 \\ \hline
         \multirow{6}{*}{\rotatebox[origin=c]{90}{\textbf{Both-}}} & InterFusion & 0.383 & 0.504 & 0.490 & 0.649 & 0.651 & 0.753 & 0.418 & 0.512 & 0.559 & \underline{0.532} & 0.520 & \underline{0.628} & 0.496 \\ 
         & GRU-GNN & 0.329 & 0.383 & 0.440 & 0.647 & 0.689 & 0.742 & 0.587 & 0.806 & 0.684 & 0.506 & 0.474 & 0.613 & 0.517 \\ 
         & DVGCRN & 0.323 & 0.305 & 0.442 & 0.615 & 0.575 & 0.744 & 0.480 & 0.612 & 0.604 & 0.397 & 0.322 & 0.554 & 0.479 \\ 
         & GraphS4mer & 0.405 & 0.567 & 0.514 & 0.638 & 0.667 & 0.738 & 0.565 & 0.762 & 0.667 & 0.524 & 0.511 & 0.621 & \underline{0.533} \\ 
         & IMDiffusion & \underline{0.426} & \underline{0.603} & \underline{0.533} & 0.611 & 0.553 & \underline{0.750} & 0.544 & 0.725 & 0.651 & 0.381 & 0.452 & 0.532 & 0.491 \\ \cdashline{2-15}  
         & \textbf{TSAD-C} & \textbf{0.479} & \textbf{0.801} & \textbf{0.604} & \textbf{0.707} & \textbf{0.841} & \textbf{0.773} & \textbf{0.652} & \textbf{0.843} & \textbf{0.728} & \textbf{0.545} & \textbf{0.830} & \textbf{0.652} & \textbf{0.596} \\ \hline 
    \end{tabular}}
    \label{tab:result_f1}
\end{table*}

\textbf{Datasets.} Many TSAD methods have relied on benchmark datasets such as Yahoo, NASA, SWaT, WADI, SMAP, and MSL. However, these datasets are unreliable due to (i) mislabeled ground truth, (ii) triviality, (iii) unrealistic anomaly density and (iv) run-to-failure bias\s{, which prompts algorithms to simply detect the last points as anomalies}  \cite{wu2021current}. This renders them unsuitable for evaluating TSAD methods. Aware of these issues, we carefully select four datasets that are  reliable, diverse, yet challenging. They are recorded using various sensor systems, each varying in the number and types of sensors, leading to distinct characteristics. These include \textbf{SMD} \cite{su2019robust}, \textbf{ICBEB} \cite{liu2018open}, \textbf{DODH} \cite{guillot2020dreem}, and \textbf{TUSZ} \cite{shah2018temple}. 

Specifically, SMD is an industrial dataset consisting of five weeks of data from 28 server machines and widely used in the TSAD field. While not perfect, SMD is considered of much higher quality compared to criticized benchmarks \cite{wagner2023timesead} (see \textbf{Appendix \ref{subsec:smd}}). ICBEB, DODH and TUSZ are well-established yet challenging datasets from the biomedical domain, have not received criticisms (i)-(iv), and importantly, they reflect real-world scenarios with diverse anomaly types, while other datasets such as SMD contain only one anomaly type. ICBEB is an ECG database, consisting of normal heart rhythms and five anomaly types. DODH is a sleep database, with the deepest sleep stage as normal and two anomaly sleep types. TUSZ is an EEG database, with normal resting-state brain activities and two anomaly seizure types. \br{Statistics of each dataset are shown in Table \ref{tab:dataset}, with further details in \textbf{Appendix \ref{sec:datasets}}}. Note that we include 10-20\% anomalies of all types in $X_{\text{train}}$ to contaminate normal data during training, increasing the challenge for algorithms to detect all anomaly types in the test phase.


\textbf{Baselines.} \br{We compare TSAD-C against 12 SOTA unsupervised methods from the TSAD literature, ranging from autoencoders, self-supervised, transformers to diffusion approaches. For a fair comparison, we do not include methods that require transfer learning with any additional datasets.} We categorize all methods into three groups based on their ability to capture either Intra-, Inter- or Both-variable dependencies. Specifically, Intra- methods include USAD \cite{audibert2020usad}, LSTM-AE \cite{wei2023lstm}, S4-AE \cite{gu2021efficiently} and DCdetector \cite{yang2023dcdetector}. Inter- methods are GAE \cite{du2022graph}, GDN \cite{deng2021graph}, and EEG-CGS \cite{ho2023self}.  Methods addressing Both- include InterFusion \cite{li2021multivariate}, DVGCRN \cite{chen2022deep}, GRU-GNN and GraphS4mer \cite{tang2023modeling}, IMDiffusion \cite{chen2024imdiffusion} and our method. Details of the baselines are shown in \br{\textbf{Appendix \ref{sec:baselines}}}.


\textbf{Evaluation Metrics.} We employ F1-score (F1), Recall (Rec), and Area Under the Precision-Recall Curve (APR) to comprehensively assess the performance of each method.


\subsection{Experimental Results}

\subsubsection{Comparison with State-of-the-Art}

The performances of all methods are presented in Table \ref{tab:result_f1}, where TSAD-C uses RandBM. It shows that TSAD-C surpasses all existing studies and achieves an average improvement of 6.3\% in F1 compared to the second-best method. TSAD-C also obtains a significant improvement in Rec -- the ability to correctly detect most of anomalies. This improvement is crucial when dealing with contaminated data, where existing methods failed to detect the anomaly types similar to those encountered during training due to their assumption of clean training data, leading to misdetection of such anomalies. Additionally, existing methods handling Both- generally outperform those addressing only one dependency type, supporting our assumption that both dependency types are crucial. For example, GraphS4mer outperforms models that focus on a single dependency type, such as USAD, which addresses only temporal aspects, or GAE, which focuses on spatial dependencies. Moreover, methods handling long-range Intra- (e.g., S4-AE) outperform those concentrating solely on Inter-. This suggests that while inter-variable dependencies are also important, they often represent variable relationships that can be captured over shorter temporal windows, which might miss the broader temporal context necessary to detect anomalies effectively.

\subsubsection{Resilience to Contamination Levels}

This section verifies TSAD-C's performance through two additional experiments on ICBEB: (1) varying the number of anomaly types and (2) varying the anomaly ratio $\eta$ in $X_{\text{train}}$ and $X_{\text{valid}}$. \br{We also select five methods, each achieving high performance within its category, for comparison.}

\begin{figure}[h]
\centering
\begin{subfigure}{0.23\textwidth}
    \includegraphics[width=\textwidth]{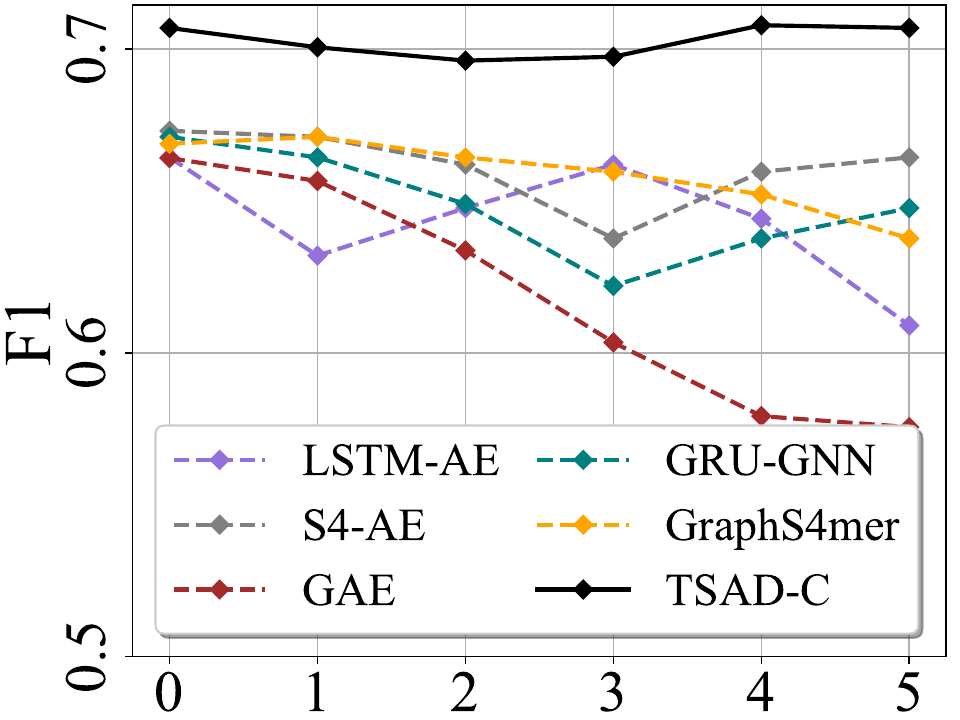}
\end{subfigure}
\begin{subfigure}{0.23\textwidth}
    \includegraphics[width=\textwidth]{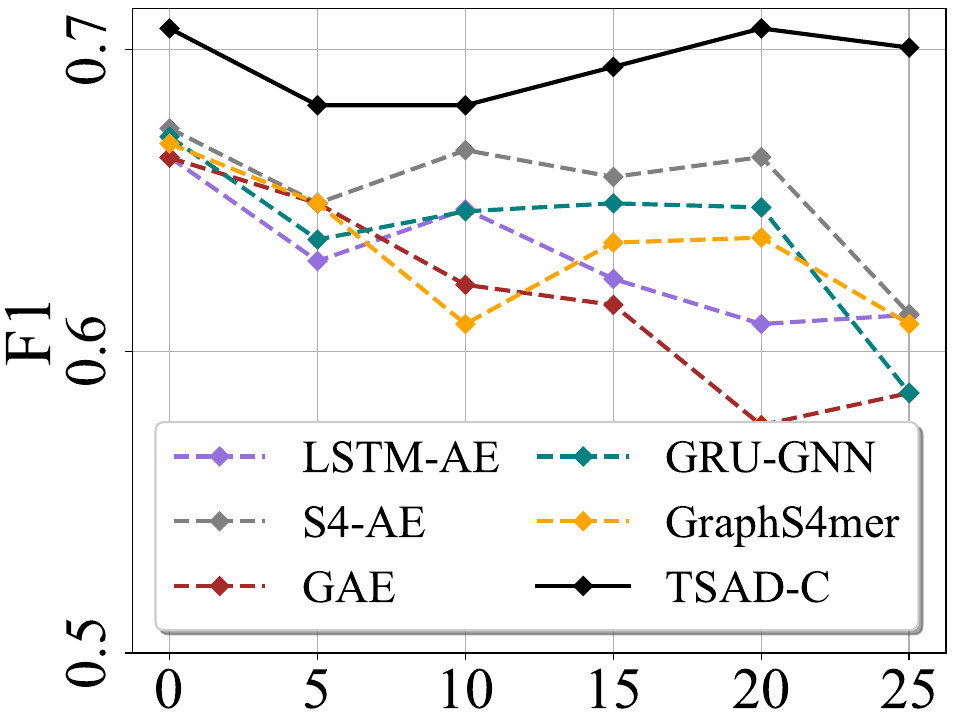}
\end{subfigure}
\caption{(\textbf{Left}) F1 score versus the number of anomaly types $\kappa$. (\textbf{Right}) F1 score versus the anomaly ratio $\eta$.}
\label{fig:sensitivity}
\end{figure}

\textbf{Variability on The Anomaly Types.} We assess TSAD-C's performance in detecting five anomaly types available in $X_{\text{test}}$ while not all types are present in $X_{\text{train}}$ and $X_{\text{valid}}$. We introduce $\kappa$ as the number of anomaly types -- e.g., $\kappa=2$ signifies two anomaly types present in $X_{\text{train}}$ and $X_{\text{valid}}$. Note that $\eta$ remains constant. Figure \ref{fig:sensitivity} (\textbf{Left}) shows that the performance of existing methods diminishes as $\kappa$ increases
since they all assume the input data as pure normal data, hence are incapable of handling contaminated data. \s{These methods face more difficulties when the data impurity increases. }
Remarkably, TSAD-C consistently attains the highest F1, irrespective of changes in $\kappa$.  This underlines our method's denoising prowess as it remains effective regardless of the diversity in available anomaly types.

\textbf{Variability on The Anomaly Ratio.} We investigate the robustness of TSAD-C by varying $\eta$. Note that all anomaly types ($\kappa=5$) are present within each subset of the dataset for this experiment. Figure \ref{fig:sensitivity} (\textbf{Right}) shows the consistent performance of TSAD-C across different anomaly ratios, demonstrating the Decontaminator's effectiveness in TSAD-C. Meanwhile, the performance of other unsupervised methods tends to decline as normal data impurity increases.

\subsubsection{Visualization of Normal Approximation}

\begin{figure}[t]
\centering
\includegraphics[width=1\linewidth]{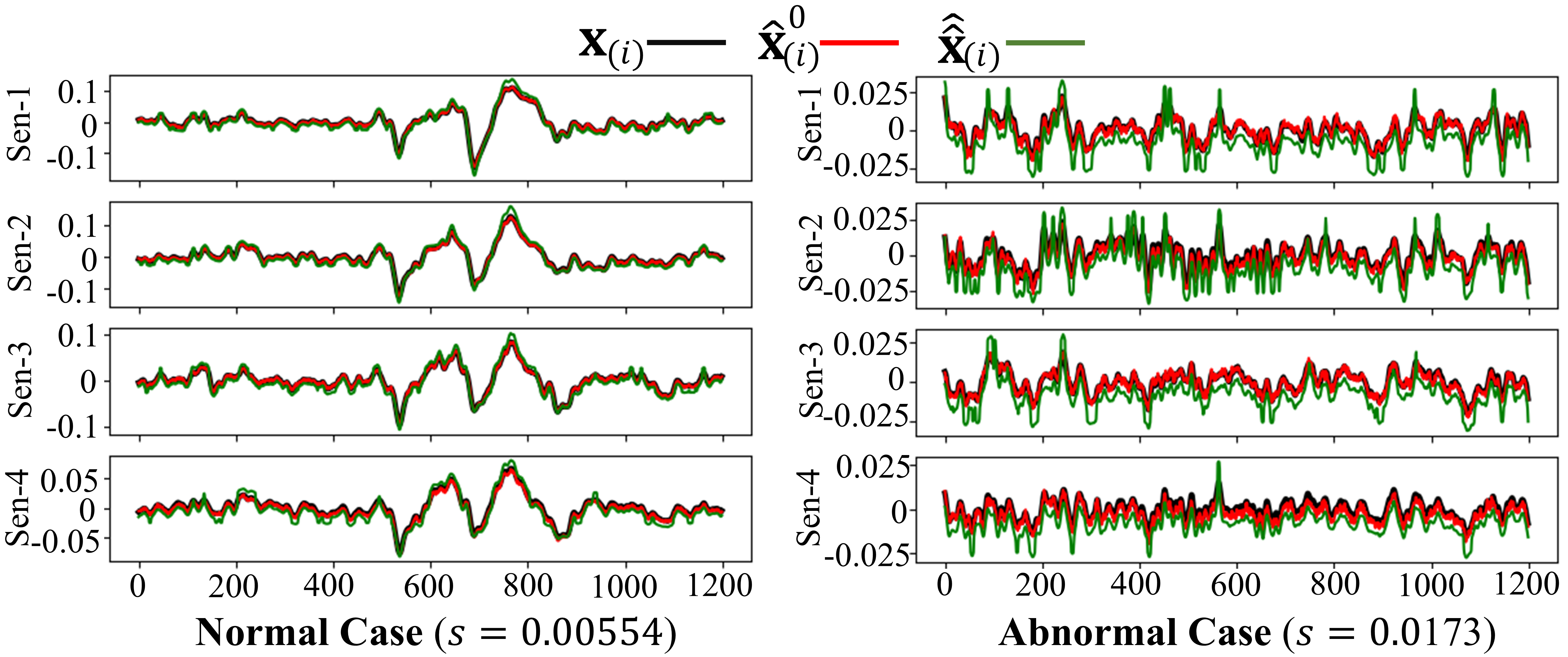}
\caption{Comparison between normal and abnormal cases for the masked segment in DODH. Sen-$k$ denotes the $k$th sensor. Each case includes $\mathbf{x}_{(i)}$, $\hat{\mathbf{x}}_{(i)}^0$ and $\mathbf{\doublehat{x}}_{(i)}$.}
\label{fig:viz_dodh}
\end{figure}

Figure \ref{fig:viz_dodh} shows a visualization comparing the ground truth $\mathbf{x}_{(i)}$, decontaminated data $\hat{\mathbf{x}}_{(i)}^0$, and reconstructed data $\mathbf{\doublehat{x}}_{(i)}$ of a masked segment of normal and abnormal samples in DODH. Visualizations on other datasets are shown in \br{\textbf{Appendix \ref{sec:add_visualizations}}}. The segment of $\mathbf{x}_{(i)}$ is masked to zeros using BoM, before being processed by the Decontaminator.
It is shown that $\hat{\mathbf{x}}_{(i)}^0$ and $\mathbf{\doublehat{x}}_{(i)}$ fit $\mathbf{x}_{(i)}$ very well in the normal case, leading to a lower $s$. Meanwhile, there are significant fluctuations of $\hat{\mathbf{x}}_{(i)}^0$ and $\mathbf{\doublehat{x}}_{(i)}$ compared to $\mathbf{x}_{(i)}$ in the abnormal case, resulting in a much higher $s$. This shows TSAD-C's effectiveness in distinguishing anomalies from normal data.

\subsubsection{Ablation Study}

Certainly, handling contaminated data requires the indispensable inclusion of the Decontaminator. Excluding it would align the TSAD-C's detection results with those of other unsupervised methods. Hence, we conduct ablation studies specifically focusing on Module (2) - Long-range Variable Dependency Modeling, and Module (3) - Anomaly Scoring, as shown in Table \ref{tab:components}. RandBM is used in all experiments. The sign ``--" denotes the exclusion of a component.

\textbf{Module (2).} The results are shown in the first and second rows, where Intra and Inter, respectively, denote the Intra-, and Inter-variable Modeling. Note that $s_2$ is computed by the non-removed block of Module (2), i.e., for the first and second rows, $s_2$ is determined by Inter and Intra, respectively. The results highlight the superior performance achieved by capturing long-range intra-variable dependencies, showing that while Inter is also an important component, having Intra is often prioritized due to the nature of time-series data.

\setlength{\tabcolsep}{2.5pt}
\renewcommand{\arraystretch}{1.5} 
\setlength{\tabcolsep}{3.1pt}
\begin{table}[!b]
\centering
\caption{The performance of individual components in TSAD-C and their combinations. M. (2) and M. (3) denote Module (2) and Module (3), respectively.}
\scalebox{0.75}{
\begin{tabular}{c c c c c c c c c c c }
    \hline 
     \multicolumn{2}{c}{\multirow{2}{*}{\textbf{TSAD-C}}} & \multicolumn{3}{c}{ICBEB} & \multicolumn{3}{c}{DODH} & \multicolumn{3}{c}{TUSZ} \\ \cmidrule(lr){3-5} \cmidrule(lr){6-8} \cmidrule(lr){9-11}
     & & F1 & Rec & APR & F1 & Rec &  APR & F1 & Rec & APR \\ \hline
     \multirow{2}{*}{\rotatebox[origin=c]{90}{M. (2)}} 
     & -- Intra & 0.651 & 0.735 & 0.738 & 0.565 & 0.637 & 0.667 & 0.499 & 0.655 & 0.598 \\ \cdashline{2-11} 
     & -- Inter & 0.686 & 0.803 & 0.759 & 0.622 & 0.762 & 0.705 & 0.535 & 0.789 & 0.639 \\ \hline
     \multirow{2}{*}{\rotatebox[origin=c]{90}{M. (3)}} 
     & -- $s_1$ & 0.706 & \textbf{0.856} & 0.771 & 0.637 & 0.80 & 0.716 & 0.517 & 0.745 & 0.622 \\ \cdashline{2-11} 
     & -- $s_2$ & 0.699 & 0.818 & 0.768 & 0.627 & 0.787 & 0.709 & 0.518 & 0.712 & 0.617 \\ \hline 
     \multicolumn{2}{c}{\textbf{All}} & \textbf{0.707} & 0.841 & \textbf{0.773} & \textbf{0.652} & \textbf{0.843} & \textbf{0.728} & \textbf{0.545} & \textbf{0.830} & \textbf{0.652} \\ \hline 
     
\end{tabular}}
\label{tab:components}
\end{table}

\textbf{Module (3).} In this study, all components of Modules (1) and (2) are available during training, whereas we exclude either $s_1$ or $s_2$, respectively, obtained from Module (1) or Module (2) during testing. The results, shown in the third and fourth rows, indicate that by removing either $s_1$ or $s_2$, we observe a small drop in performance. This is due to the fact that in the test phase, we still perform data decontamination using the Decontaminator trained during training. Such data decontamination during testing resembles a case where one performs the unsupervised anomaly detection of the clean test data. This experiment can be considered as a demonstration of the Decontaminator's effectiveness.
%
Importantly, integrating all components together (aka \textbf{All}) yields the best performance across all datasets, showcasing the synergistic complementarity of each component in enhancing the model's ability to detect anomalies.

\subsubsection{Decontaminator Efficiency Study} \label{subsec:decontaminator_study}

As described in Section~\ref{subsec:decontaminator}, we estimate the noise exclusively on the masked portions using $\mathcal{L}_{\text{noise}}$, shown in Equation~\eqref{eq: loss_noise}. This approach simplifies and streamlines the reverse process and can be accomplished using less intricate networks. To empirically demonstrate this, we conduct additional experiments, where the noise is minimized over the entire observation (rather than just the masked parts), as commonly done in diffusion studies \cite{chen2024imdiffusion}. This loss is defined as $ \mathcal{L}_{\text{noise}}^{\text{entire}} = \|\epsilon_t - \hat{\epsilon}_t \|^2$. The results, shown in Table~\ref{tab:compare_loss}, indicate that minimizing  $\mathcal{L}_{\text{noise}}$ exclusively on the masked portions achieves better performance, and is more applicable as the training time, is much faster, hence the decontaminated data can be quickly prepared for the second module. Meanwhile, minimizing $\mathcal{L}_{\text{noise}}^{\text{entire}}$ achieves suboptimal performance. We observe that this could be improved by increasing the complexity of the network, e.g., increasing the number of the network layers and the number of reverse steps. However, it would result in the slower training speed, which is not desirable for the Decontaminator in our task.

\renewcommand{\arraystretch}{1.35}
\setlength{\tabcolsep}{2.5pt}
\begin{table}[t]
\centering
\caption{\br{Comparison of minimizing $\mathcal{L}_{\text{noise}}^{\text{entire}}$ versus $\mathcal{L}_{\text{noise}}$. All components of TSAD-C are included in the training phase, whereas in the test phase, either $s_1$ (i.e., $-s_2$), $s_2$ (i.e., $-s_1$), or both $s_1$ and $s_2$ (i.e., $s_1+s_2$) are included (min: minutes).}}
\scalebox{0.76}{
\begin{tabular}{c c c c c c c c} 
     \hline
      \multirow{3}{*}{\rotatebox[origin=c]{90}{\textbf{Dataset}}} & \multirow{3}{*}{\textbf{TSAD-C}} & \multicolumn{3}{c}{$\mathcal{L}_{\text{noise}}^{\text{entire}}$} & \multicolumn{3}{c}{$\mathcal{L}_{\text{noise}}$}  \\ \cmidrule(lr){3-5} \cmidrule(lr){6-8}
     & & \makecell{Training time \\ per epoch (min)}  & \multicolumn{1}{c}{F1} & \multicolumn{1}{c}{APR} & \makecell{Training time \\ per epoch (min)} & \multicolumn{1}{c}{F1} & \multicolumn{1}{c}{APR} \\ \hline 
     \multirow{3}{*}{\rotatebox[origin=c]{90}{ICBEB}} & -- $s_1$ & 14.6 & 0.590 & 0.709 & 12.2 & 0.706 & 0.771  \\ \cdashline{2-8} 
     & -- $s_2$ & 14.6 & 0.686 & 0.759 & 12.2 & 0.699 & 0.768 \\ \cdashline{2-8} 
     & $s_1 + s_2$ & 14.6 & 0.679  & 0.756 & 12.2 & \textbf{0.707} & \textbf{0.773}  \\ \hline
     \multirow{3}{*}{\rotatebox[origin=c]{90}{DODH}} & -- $s_1$ & 21.8 & 0.617 & 0.702 & 19.3 & 0.637 & 0.716  \\ \cdashline{2-8} 
     & -- $s_2$ & 21.8 & 0.600 & 0.703 & 19.3 & 0.627 & 0.709  \\ \cdashline{2-8} 
     & $s_1 + s_2$ & 21.8 & 0.623 & 0.700 & 19.3 & \textbf{0.652} & \textbf{0.728}   \\ \hline
     \multirow{3}{*}{\rotatebox[origin=c]{90}{TUSZ}} & -- $s_1$ & 40.5 & 0.498 & 0.597 & 38.4 & 0.517 & 0.622  \\ \cdashline{2-8} 
     & -- $s_2$ & 40.5 & 0.481 & 0.580 & 38.4 & 0.518 & 0.617 \\ \cdashline{2-8} 
     & $s_1 + s_2$ & 40.5 & 0.505 & 0.603 & 38.4 & \textbf{0.545} & \textbf{0.652}  \\ \hline
\end{tabular}}
\label{tab:compare_loss}
\end{table}

\subsubsection{Effect of Masking Strategy}

This section evaluates the performance of TSAD-C with various masking strategies. Note that we maintain a fixed value for $r$, which is used to control the masking ratio, to avoid hyperparameter fine-tuning. The consistency of better performance of TSAD-C even with a fixed $r$, irrespective of changes in the anomaly ratio $\eta$, is demonstrated in Figure \ref{fig:sensitivity} (\textbf{Right}). Table \ref{tab:compare_masking} shows that the performance of TSAD-C remains largely consistent when employing either RandM, RandBM, or BoM, \br{across varying $\kappa$ present in  $X_\text{train}$ and $X_\text{valid}$}. Despite the consistency, RandBM showcases the most optimal performance. This can be attributed to introducing randomness into the masked time windows, which increases the level of robustness and adaptability to diverse patterns in the data. This is crucial for real-world applications where anomalies can manifest in diverse ways across sensors. RandBM emulates this diversity, enabling the model to learn and detect sensor-specific anomalies more effectively, thus boosting its overall performance.

\renewcommand{\arraystretch}{1.4}
\setlength{\tabcolsep}{3.5pt}
\begin{table}[t]
\centering
\caption{\br{Comparison of masking strategies employed in TSAD-C across varying numbers of anomaly types $\kappa$ in $X_\text{train}$ and $X_\text{valid}$. SD stands for standard deviation.}}
\scalebox{0.75}{
\begin{tabular}{c c c c c c c c} 
     \hline
      \multirow{2}{*}{\rotatebox[origin=c]{90}{\textbf{Dataset}}} & \multirow{2}{*}{\makecell{\textbf{Anomaly } \\ \textbf{Type} \\ ($\kappa$)}} & \multicolumn{2}{c}{RandM} & \multicolumn{2}{c}{RandBM} & \multicolumn{2}{c}{BoM}  \\ \cmidrule(lr){3-4} \cmidrule(lr){5-6} \cmidrule(lr){7-8} 
      & & F1 & APR & F1 & APR & F1 & APR \\ \hline 
     \multirow{7}{*}{\rotatebox[origin=c]{90}{ICBEB}} & 0 & 0.702 & 0.771 & 0.707 & 0.773 & 0.694 & 0.764  \\  \cdashline{2-8}
     & 1 & 0.669 & 0.748 & 0.687 & 0.759 & 0.671 & 0.752  \\  \cdashline{2-8}
     & 2 & 0.701 & 0.768 & 0.696 & 0.742 & 0.661 & 0.743  \\ \cdashline{2-8}
     & 3 & 0.681 & 0.755 & 0.697 & 0.751 & 0.675 & 0.751  \\ \cdashline{2-8}
     & 4 & 0.688 & 0.749 & 0.708 & 0.759 & 0.662 & 0.745  \\ \cdashline{2-8}
     & 5 & 0.694 & 0.764 & 0.707 & 0.773 & 0.669 & 0.747  \\ \cdashline{2-8} 
     & \makecell{Mean \\ $\pm$ SD} & \makecell{0.689 \\ $\pm$ 0.012} & \makecell{0.759 \\ $\pm$ 0.009} & \makecell{\textbf{0.700} \\ $\pm$ 0.007} & \makecell{\textbf{0.760} \\ $\pm$ 0.011} & \makecell{0.672 \\ $\pm$ 0.011} & \makecell{0.750 \\ $\pm$ 0.007}  \\ \hline
     \multirow{4}{*}{\rotatebox[origin=c]{90}{DODH}} & 0 & 0.662 & 0.735 & 0.632 & 0.713 & 0.643 & 0.722  \\  \cdashline{2-8}
     & 1 & 0.637 & 0.717 & 0.671 & 0.742 & 0.618 & 0.703  \\ \cdashline{2-8}
     & 2 & 0.647 & 0.724 & 0.652 & 0.728 & 0.623 & 0.707  \\ \cdashline{2-8} \cdashline{2-8}
     & \makecell{Mean \\ $\pm$ SD} & \makecell{0.649 \\ $\pm$ 0.010} & \makecell{0.725 \\ $\pm$ 0.007} & \makecell{\textbf{0.651} \\ $\pm$ 0.016} & \makecell{\textbf{0.727} \\ $\pm$ 0.012} & \makecell{0.628 \\ $\pm$ 0.011} & \makecell{0.711 \\ $\pm$ 0.008}  \\ \hline
     \multirow{4}{*}{\rotatebox[origin=c]{90}{TUSZ}} & 0 & 0.541 & 0.649 & 0.530 & 0.639 & 0.531 & 0.640  \\  \cdashline{2-8}
     & 1 & 0.527 & 0.632 & 0.543 & 0.645 & 0.520 & 0.623  \\ \cdashline{2-8}
     & 2 & 0.519 & 0.631 & 0.545 & 0.652 & 0.531 & 0.640  \\    \cdashline{2-8}
     & \makecell{Mean \\ $\pm$ SD} & \makecell{0.529 \\ $\pm$ 0.009} & \makecell{0.637 \\ $\pm$ 0.008} & \makecell{\textbf{0.539} \\ $\pm$ 0.006} & \makecell{\textbf{0.645} \\ $\pm$ 0.005} & \makecell{0.527 \\ $\pm$ 0.005} & \makecell{0.634 \\ $\pm$ 0.008}  \\ \hline
\end{tabular}}
\label{tab:compare_masking}
\end{table}

\section{Conclusion}
This paper introduces TSAD-C, the first method trained on contaminated data to detect all types of anomalies in multivariate time series. TSAD-C comprises the Decontaminator, aimed at removing the potential anomalies during training and swiftly preparing decontaminated data for subsequent modules. The Long-range Variable Dependency Modeling module is designed to capture long-range intra- and inter-variable dependencies, and provide an approximation of purified data. The Anomaly Scoring module integrates the capability of the first two modules. We demonstrate the superior performance of TSAD-C on four reliable, diverse and challenging datasets compared to existing methods.


\bibliography{uai2025-template}

\newpage

\onecolumn

\title{Contaminated Multivariate Time-Series Anomaly Detection with Spatio-Temporal Graph Conditional Diffusion Models\\(Appendix)}
\maketitle

\appendix

\section{Related Work} \label{sec:related_works}

\subsection{Time-Series Anomaly Detection Techniques}

Existing TSAD methods can be broadly categorized into supervised and unsupervised approaches. Early supervised methods require complete anomaly-labeled datasets that encompass the complete distribution of all potential anomalies during training \cite{carmona2022neural,tang2021self,tang2023modeling}. However, acquiring labels for all possible anomalies is  impractical for real-world applications. As a result, \t{a recent work} \cite{lai2023open} has explored TSAD in open-set scenarios \cite{ding2022catching,tian2022anomaly}, where only a limited amount of labeled anomalous data is available during training. Although this reduces the dependency on complete anomaly-labeled data, open-set approaches still require a small number of labeled anomalies during training. Nonetheless, obtaining even a few labeled anomalies during data collection can be challenging in many cases. For example, epileptic seizures in brain recordings may occur unpredictably, often requiring prolonged monitoring to capture seizure events. Similarly, anomalies in industrial machinery, such as machine failures, are rare and might only occur after extended periods of operation.


\br{To address these challenges, many studies have proposed unsupervised methods, assuming that only normal data is available during the training phase \cite{su2019robust,audibert2020usad,shen2020timeseries,deng2021graph,li2021multivariate,chen2022deep,yang2023dcdetector,ho2023self,hojjati2023multivariate,chen2024imdiffusion}. Recent unsupervised methods, mostly leveraging deep learning architectures, can be further divided into four main categories based on the employed loss functions minimized during training: Autoencoder (AE)-based, Generative Adversarial Network (GAN)-based, Predictive-based, and Self-supervised methods  \cite{ho2023graph}. Specifically, AE-based methods utilize reconstruction loss to minimize the difference between the input data and its reconstructed output within an encoder-decoder framework \cite{su2019robust,chen2022deep}. GAN-based methods leverage the losses from both the generator and discriminator during training \cite{liang2021consistent,deng2022graph}. Predictive-based methods utilize prediction error, aiming to forecast future values or properties rather than simply reconstructing the input data \cite{zhao2020multivariate,deng2021graph,han2022learning}. Lastly, self-supervised methods employ contrastive loss or other pretext-task-based losses to learn meaningful representations from the training data \cite{ho2023self,hojjati2023multivariate}. However, all these categories of unsupervised methods assume that the training data must be clean and do not address the challenges posed by contaminated data, limiting their effectiveness in accurately detecting anomalies in practical scenarios and applications.}

\subsection{Diffusion Models}

\br{Diffusion models, a class of generative models \cite{sohl2015deep,croitoru2023diffusion}, have gained significant attention for anomaly detection across various domains. For example, \cite{livernoche2023diffusion} introduced diffusion models to capture normal data distributions, enabling anomaly detection in tabular data. In image anomaly detection, \cite{wolleb2022diffusion} proposed a diffusion model that demonstrates superior performance compared to the baseline methods, by generating normal image patterns, with anomalies detected as deviations from these patterns. In video anomaly detection, \cite{flaborea2023multimodal} proposed a multimodal approach with diffusion models to detect anomalies in video data. Additionally, \cite{karami2024graph} proposed a diffusion model coupled with a graph-based module for modeling intra- and inter-variable dependencies existing in video data, to effectively detect anomalies. }

\br{In recent years, researchers have begun to explore the potential of diffusion models in the \t{signal} domain. For example, \cite{tashiro2021csdi} leveraged diffusion models for time-series imputation tasks, demonstrating strong performance in handling missing data. Likewise, \cite{alcaraz2022diffusion,rasul2021autoregressive} applied diffusion models for both time-series imputation and forecasting tasks, showcasing their capability in achieving accurate future predictions. Notably, \cite{chen2024imdiffusion} marked the first to adapt diffusion models specifically for TSAD tasks. \cite{chen2024imdiffusion} considered diffusion models as time-series imputers to fill in missing data while simultaneously detecting anomalies. Despite these advancements, no existing studies have explored the use of diffusion models in the context of contaminated time-series data, where anomalies sneak into the normal training data. Addressing this challenge would unlock new potential for diffusion models, enabling them to tackle more complex real-world scenarios and applications.}

\subsection{Long-term Intra- and Inter-variable Dependency Modeling}

\br{In time-series data, capturing both long-range intra- and inter-variable dependencies is essential for effective anomaly detection\t{\cite{ho2023graph}}. Specifically, capturing long-range intra-variable dependencies is valuable for TSAD, as they help differentiate between normal and abnormal temporal variations. However, many existing methods overlook this critical aspect by preprocessing data into short-term intervals \cite{carmona2022neural,lai2023open,chen2024imdiffusion} or by relying on conventional temporal networks, such as recurrent neural networks \cite{dai2022graph,hojjati2023multivariate} and transformers \cite{yang2023dcdetector,chen2024imdiffusion}.}
%
\t{Such data preprocessing steps may discard important information encoded in raw signals and
neglect long-range intra-variable dependencies \cite{tang2023modeling}. Moreover, conventional temporal models struggle in handling sequences that are sampled at high sampling rates, leading to long sequences that can be up to thousands of time steps \cite{tay2020long}.}
%
%
\br{Recently, advanced deep sequence models, such as structured state space models \cite{gu2021efficiently}, have demonstrated superior performance in long sequence modeling tasks, outperforming previous state-of-the-art models in speech classification \cite{gu2021efficiently}, audio generation \cite{goel2022s}, and time-series forecasting \cite{alcaraz2022diffusion}. Despite these successes, the potential of these models in unsupervised TSAD tasks remains unexplored. This presents an exciting opportunity for TSAD research to improve anomaly detection in complex and long-range time-series data.}

\br{Additionally, many recent studies have leveraged graphs to model inter-variable dependencies effectively. This property is valuable for TSAD as the changes of one variable can be used to predict the changes of another variable if they are related. For instance, several studies have proposed constructing static graphs based on prior knowledge, such as the Euclidean distances between sensors \cite{tang2021self,tang2023modeling}, the correlation matrices of sensor features \cite{ho2023self}, or the mutual information between sensor features \cite{hojjati2023multivariate} in multivariate sensory systems. However, this approach is often impractical for many real-world applications due to the dynamic nature of testing environments, e.g., in electroencephalogram monitoring of epilepsy patients, sensor placements may vary} \t{between subjects depending on many factors such as scalp injuries, epilepsy types and targeted brain regions. This results in variations in the inter-variable relationships across subjects, making it challenging for models employing static graphs to generalize effectively to diverse testing scenarios.} 
%
Hence, dynamically learning the graphs is highly desirable as this approach allows for capturing the inter-variable dependencies over time \cite{dai2022graph,deng2021graph,han2022learning}, leading to improved anomaly detection performance in complex systems.

\section{The pseudocodes of TSAD-C} \label{sec:pseudocodes}

\br{To clearly outline the framework of TSAD-C and facilitate reproducibility, we present the pseudocode for the training phase, as well as the test phase of TSAD-C in Algorithm~\ref{alg:alg_train} and Algorithm~\ref{alg:alg_test}, respectively.}

\begin{algorithm}[H]
\small
\caption{Training Phase of TSAD-C}
\label{alg:alg_train}
\textbf{Input}: $X_{\text{train}}$, $X_{\text{valid}}$, $\mathbf{v}$, hyperparameters $\{\beta_0,\beta_T,T\}$, $\xi_1$, $\xi_2$, $\xi_3$. \\
\textbf{Output}: Learnable parameters $\Theta$.
\begin{algorithmic}[1] 
\STATE Compute $\mathbf{x}_{(i)}^u = \mathbf{x}_{(i)} \odot \mathbf{v}$, where $ \mathbf{x}_{(i)} \in X_{\text{train}}$. 
\STATE Compute diffusion hyperparameters $\Bar{\alpha}_t$, $\Bar{\beta}_t$, $\epsilon_t$ and $c$.
\STATE Compute $\mathbf{x}_{(i)}^t$.
\STATE Randomly initialize learnable parameters $\Theta$.\\
\WHILE{\textit{not converged}}
    \STATE \textit{// Decontaminator} 
    \STATE  Compute $\hat{\epsilon}_t = \epsilon_{\theta}(\mathbf{x}_{(i)}^t,t,c)$. \\
    \STATE $\mathcal{L}_{\text{noise}} = \|\epsilon_t \odot (1-\mathbf{v})- \hat{\epsilon}_t \odot (1-\mathbf{v}) \|^2$. \\
    \STATE Compute $\hat{\mathbf{x}}_{(i)}^0$.
    \STATE \textit{// Long-range Variable Dependency Modeling}
    \STATE Compute $\mathbf{H}_{(i)} = \text{S4}(\hat{\mathbf{x}}_{(i)}^0)$.
    \STATE Construct $\mathbb{G}_{(i)} = \{\mathcal{G}_{(i)}^m\}_{m=1}^d$.
    \STATE $\mathcal{L}_{\text{graph}} = 0$.
    \FOR{$m \in 1,2,...,d$}
    \STATE Construct $\mathcal{G}_{(i)}^m = \{\mathbf{E}_{(i)}^m, \mathcal{A}_{(i)}^m\}$, where $\mathbf{E}_{(i)}^m = \text{mean-pool}(\mathbf{H}_{(i)})$, $\mathcal{A}_{(i)}^m = \text{Graph-Learning}(\mathbf{E}_{(i)}^m,\mathcal{A'}_{(i)}^{m})$. \\
    \STATE $\mathcal{L}_{\text{graph}} = \mathcal{L}_{\text{graph}} +  \xi_1 \mathcal{L}_{\text{smooth}} (\textbf{E}_{(i)}^m, \mathcal{A}_{(i)}^m) +
        \xi_2 \mathcal{L}_{\text{sparse}}(\mathcal{A}_{(i)}^m)  + \xi_3 \mathcal{L}_{\text{connect}}(\mathcal{A}_{(i)}^m)$.
    \STATE Compute $\mathbf{z}_{(i)}^{m} = \text{GIN} (\mathbf{E}_{(i)}^m, \mathcal{A}_{(i)}^m)$. \\
    \ENDFOR
    \STATE Compute $\mathbf{Z}_{(i)} = \text{concat} (\mathbf{z}_{(i)}^1, \ldots, \mathbf{z}_{(i)}^d)$. \\
    \STATE Compute $\doublehat{\mathbf{x}}_{(i)} = \text{Linear} (\mathbf{Z}_{(i)})$.
    \STATE $\mathcal{L}_{\text{recon}} = \|\hat{\mathbf{x}}_{(i)}^0 - \doublehat{\mathbf{x}}_{(i)}\|^2.$ \\
    \STATE $\mathcal{L} = \mathcal{L}_{\text{noise}} + \mathcal{L}_{\text{graph}} + \mathcal{L}_{\text{recon}}.$
    \STATE Backpropagate $\mathcal{L}$ to update $\Theta$.
    \STATE Early stopping using $X_{\text{Valid}}$.
\ENDWHILE
 \\
\end{algorithmic}
\end{algorithm}

\begin{algorithm}[H]
\small
\caption{Inference Phase of TSAD-C}
\label{alg:alg_test}
\textbf{Input}: $X_{\text{test}}$, $X_{\text{valid}}$, $\mathbf{v}$, hyperparameters $\{\beta_0,\beta_T,T\}$, $\lambda_1$, $\lambda_2$.\\
\textbf{Output}: Anomaly score $s$.
\begin{algorithmic}[1] 
\STATE \textit{// Decontaminator} 
\STATE Compute $\mathbf{x}_{(i)}^u = \mathbf{x}_{(i)} \odot \mathbf{v}$, where $ \mathbf{x}_{(i)} \in X_{\text{test}}$. 
\STATE Compute diffusion hyperparameters $\Bar{\alpha}_t$, $\Bar{\beta}_t$, $\epsilon_t$ and $c$.
\STATE Compute $\mathbf{x}_{(i)}^T$.
\FOR{$t = T, T-1, \ldots, 1$}
\STATE Compute $\mu_{\theta}(\mathbf{x}_{(i)}^t,t,c)$ and $\sigma_{\theta}(\mathbf{x}_{(i)}^t,t,c)$.
\STATE Sample $\mathbf{x}_{(i)}^{t-1} \sim q_{\theta}(\mathbf{x}_{(i)}^{t-1}|\mathbf{x}_{(i)}^t) \coloneqq$ $\mathcal{N}(\mathbf{x}_{(i)}^{t-1}; \mu_{\theta}(\mathbf{x}_{(i)}^t,t,c),$  $\sigma_{\theta}(\mathbf{x}_{(i)}^t,t,c)^2\mathbf{I})$.
\ENDFOR
\RETURN $\hat{\mathbf{x}}_{(i)}^0$.
\STATE \textit{// Long-range Variable Dependency Modeling}
\STATE Compute $\mathbf{H}_{(i)}$, $\mathbb{G}_{(i)}$, $\mathbf{Z}_{(i)}$.
\STATE Compute $\doublehat{\mathbf{x}}_{(i)}$.
\STATE \textit{// Anomaly Scoring} 
\STATE Compute $s_1 = \text{RMSE}(\hat{\mathbf{x}}_{(i)}^0, \mathbf{x}_{(i)})$.
\STATE Compute $s_2 = \text{RMSE}(\doublehat{\mathbf{x}}_{(i)}, \mathbf{x}_{(i)})$.
\STATE Compute $s = \lambda_1 s_1 + \lambda_2 s_2$.
\STATE Search $\tau$ based on the unlabeled $X_{\text{valid}}$.
\IF{$s > \tau$}
\STATE $\mathbf{x}_{(i)}$ is an anomaly.
\ELSE 
\STATE $\mathbf{x}_{(i)}$ is a normal observation.
\ENDIF
\end{algorithmic}
\end{algorithm}

\section{Datasets}\label{sec:datasets} 

In this section, we describe four reliable datasets collected from various domains that are used in our study, including SMD, ICBEB, DODH and TUSZ datasets.

\subsection{SMD} \label{subsec:smd}

The Server Machine Dataset (SMD) \cite{su2019robust} is a multivariate time-series dataset comprising five weeks of data from 28 different server machines collected by a large Internet company. SMD is widely used in the TSAD field and is considered of much higher quality \cite{wagner2023timesead} compared to other benchmarks such as Yahoo \cite{laptev2015s5}, NASA \cite{hundman2018detecting}, SWaT \cite{mathur2016swat}, WADI \cite{ahmed2017wadi}, SMAP \cite{hundman2018detecting}, and MSL \cite{hundman2018detecting}. However, several concerns have been raised about this dataset. First, \cite{wu2021current} highlighted the triviality issue in SMD, noting that the dataset can be easily solved by just one line of code. Second, according to \cite{wagner2023timesead}, SMD exhibits a distributional shift issue, where there is a shift in the statistical distribution between the training and test sets. The authors identified and excluded several machines from the dataset upon observing significant changes in mean and standard deviation between the training and test sets for those machines. They then asserted that their modified dataset is more suitable for benchmarking TSAD algorithms.

Aware of these issues, we implement data preprocessing steps that differ significantly from previous studies \cite{su2019robust,li2021multivariate,chen2022deep,yang2023dcdetector,chen2024imdiffusion} to address those concerns. First, we divide the dataset into observations (intervals), each with a length of 600. We then perform interval-level anomaly detection, as defined in our problem definition in \textbf{Section 2} - Paragraph 1 of the main paper, aligning with recent standards in TSAD \cite{carmona2022neural,lai2023open}. This approach contrasts with point-level anomaly detection, which introduces point-adjustment bias -- a common issue in the TSAD field \cite{kim2022towards}, i.e., without the point-adjustment strategy, the performance of existing methods cannot even outperforms a random baseline. 

Second, we implicitly and partially mitigate the triviality issue by treating the time series as a multivariate series, emphasizing the importance of capturing inter-variable dependencies, rather than focusing solely on individual sensors as commonly done in previous studies \cite{su2019robust,yang2023dcdetector}. To build a robust and generalized method, we also combine all 28 machines and train only a single model for both our proposed
method and compared methods. Note that, previous studies \cite{su2019robust} suggested that the data from each
of the 28 machines should be trained and tested separately. Additionally, by adopting interval-level anomaly detection instead of point-level anomaly detection, we increase the complexity and challenge for TSAD algorithms. Lastly, to address the distributional shift issue, we follow the procedure outlined in \cite{wagner2023timesead}, and examine the mean and standard deviation of the training and test sets to minimize any shift in the dataset.

Given these crucial procedures, we ensure a more rigorous and robust evaluation of TSAD algorithms on SMD. Hence, the results of our implementation on compared methods may differ from those originally reported in their papers.

\subsection{ICBEB}

The International Conference on Biomedical Engineering and Biotechnology (ICBEB) \cite{liu2018open} is an
electrocardiogram (ECG) dataset. It consists of normal data and different types of abnormal cardiac disorders. Each recording is annotated by up to three ECG experts and  might be associated with more than one abnormal classes. We select the normal ECG waveforms as normal data, while five abnormal types, including atrial fibrillation, first-degree atrioventricular, right bundle branch, premature ventricular contraction and ST-segment elevation, are selected as abnormal data. We follow the preprocessing steps as described in \cite{strodthoff2020deep,tang2023modeling}. The sampling frequency rate is 100Hz, resulting in observations with the length of 6,000 time steps. Each observation consists of 12 sensors.  Our task is to detect abnormal cardiac observations from all five types that deviate from the normal cardiac activities. 

\subsection{DODH}

The Dreem Open Dataset Healthy, called DODH \cite{guillot2020dreem} is a sleep dataset collected from polysomnographic (PSG) signals of 25 volunteers.  Each recording includes different sleep stages, namely Awake, rapid eye movement (REM), and non-REM sleep stage N3, and is annotated by a consensus of five experienced PSG readers. We select N3 (the deepest sleeping stage) as normal data since the brain activity during this stage has an identifiable pattern of delta waves; and body activities such as breathing and muscle tone rate decrease. Meanwhile, REM and Awake stages are abnormal since the brain activity during these periods rises up; and the body activities such as the eyes and the muscles start increasing. The PSG signals are recorded from different locations across the body (brain-eyes-heart-legs), presenting a very challenging problem for TSAD algorithms due to inconsistent signal patterns. There is a total of 16 recording sensors: 12 EEG sensors, 1 electromyographic (EMG) sensor, 2 electrooculography (EOG) sensors, and 1 ECG sensor. Following the preprocessing steps described in \cite{tang2023modeling}, we resample PSG signals with a sampling frequency rate of 250Hz, leading to a length of 7,500 time steps for every observation. Our task is to detect abnormal sleep observations from two types that deviate from the normal sleep activities.

\subsection{TUSZ}

The University Hospital Seizure Detection Corpus, called TUSZ \cite{shah2018temple} is the largest public electroencephalogram (EEG) seizure dataset to date. It contains 5612 EEG files with 3050 annotated seizures, and different types of seizures are defined although some of them are less presented. Annotations are made through a consensus of at least three experienced EEG readers. For each patient, there are several sessions that consist of files related to one or more recordings. For each recording, there are five EDF files containing the raw EEG signals and a header indicating frequency, duration and date. In our experiments, we select resting states of the brain activities as normal data, while abnormal data is collected from two types of seizures (focal and generalized).  Each EEG recording consists of 19 sensors. Note that compared to other types of signals such as ECG that represents fairly predictable heart waveforms, TUSZ contains much more stochastic signals due to various physiological states in the brain. Following prior studies on seizure analysis \cite{tang2021self,ho2023self}, we resample all EEG signals with a sampling frequency rate of 200Hz, resulting in a length of 12,000 time steps for each observation. Our task is to detect whether there exist abnormal patterns from two types of seizures in the observation.

It is important to re-emphasize that \textbf{ICBEB}, \textbf{DODH} and \textbf{TUSZ} are well-established yet challenging datasets and have not received criticisms in the TSAD field. 
To minimize human labeling errors, they are annotated by a panel of 3-5 experts, who have practical experience dealing with patients in clinical settings, and take into account factors such as patient history, symptoms, diagnostic tests, and treatment outcomes when identifying anomalies. Additionally, they often exhibit long-range intra-variable dependencies, reflecting phenomena like physiological states or disease progression, which oversimplified models cannot capture complex time-series dynamics. Importantly, unlike existing datasets such as \textbf{SMD} that contain a single anomaly type, these three datasets reflect real-world scenarios with diverse anomaly types, where each type may have distinct characteristics or exhibit patterns similar to normal variability.

Note that for all datasets, we first split the data at the machine/patient level into three subsets: $X_\text{train}$, $X_\text{valid}$, and $X_\text{test}$. This means each machine/patient's data is entirely contained within one of these subsets to ensure independence between them. Within each machine/patient, each observation is collected using a non-overlapping split to avoid data leakage.

\section{Baselines} \label{sec:baselines}

As mentioned in the main paper, we compare TSAD-C with 12 SOTA unsupervised methods in the TSAD field. These methods can be categorized into three groups as follows:

\begin{itemize}
    \item Methods capturing Intra-variable dependencies include:
    \begin{itemize}
        \item \textbf{USAD} \cite{su2019robust} is a Generative Adversarial Network consisting of two Autoencoders. It is trained in a two-phase process: the first phase reconstructs the normal data, while the second phase distinguishes the real data from the data coming from the first Autoencoder. The anomaly score is a combination of the scores obtained from both phases.     
        \item  \textbf{LSTM-AE} \cite{wei2023lstm} is a Long Short-Term Memory based Autoencoder. The anomaly score is the reconstruction error.
        \item \textbf{S4-AE} \cite{gu2021efficiently} is a Structured State Space based Autoencoder. The anomaly score is the reconstruction error.
        \item \textbf{DCdetector} \cite{yang2023dcdetector} is a Transformer based framework consisting of two self-attention networks. The first is for a wider patch-wise view and the another is for a finer in-patch view. The Kullback-Liebler contrastive loss minimizes the distance between the two views for normal data, with the assumption that the views will be different for abnormal data. The anomaly score is computed based on the contrastive loss.
    \end{itemize}

    \item Methods capturing Inter-variable dependencies include:
    \begin{itemize}
        \item \textbf{GAE} \cite{du2022graph} is a Graph-based Autoencoder that learns the graph structure and features during training. The reconstruction error is used as the anomaly score.
        \item \textbf{GDN} \cite{deng2021graph} is a Predictive-based approach that employs two modules: a graph structure learning module to capture the sensor relationships and a graph attention-based forecasting module to predict future values of every sensor. The anomaly score is computed by the prediction error via a graph deviation scoring module.
        \item \textbf{EEG-CGS} \cite{ho2023self} is a Self-supervised approach that first constructs graphs based on the correlations between sensors. It then leverages local structural and contextual information within the graphs to generate positive and negative sub-graphs. EEG-CGS is trained by minimizing contrastive and generative losses, with the anomaly score derived from a combination of these losses.
    \end{itemize}
    
    \item Methods capturing Both-variable dependencies include:
    \begin{itemize}
        \item \textbf{InterFusion} \cite{li2021multivariate} is a Variational Autoencoder based approach with two stochastic latent variables, each of which learns low-dimensional inter- or intra-variable embeddings within the normal data. The reconstruction error is considered as the anomaly score.
        \item \textbf{DVGCRN} \cite{chen2022deep} is a Variational Autoencoder based approach designed to capture multilevel intra- and inter-variable information within the raw data, and the inter-variable information in the latent space. It comprises two main components to model these relationships: a forecasting model for single-time step prediction, and a reconstruction model for reconstructing the input data. The anomaly score is a combination of prediction and reconstruction errors.
        \item \textbf{GRU-GNN}  \cite{tang2023modeling} incorporates a Gate Recurrent Unit network and a Graph Neural Network model to extract both intra-variable inter-variable dependencies in time-series data. The reconstruction score is used as the anomaly score for each observation.
        \item  \textbf{GraphS4mer} \cite{tang2023modeling} is a Transformer based approach that incorporates S4 layers to extract long-range intra-variable dependencies, a graph structure learning technique based on an attention mechanism to construct graphs and Graph Neural Network layers to capture the inter-variable dependencies. The reconstruction score is used as the anomaly score.
        \item \textbf{IMDiffusion} \cite{chen2024imdiffusion} proposes a Diffusion model as a time-series imputer to capture intra- and inter-variable dependencies. It employs a Transformer network for noise estimation. In the test phase, ImDiffusion utilizes the prediction error at each denoising step and ensembles them using a voting function to determine the final anomaly scores. 
    \end{itemize}
\end{itemize}

Note that none of the existing methods incorporate a mechanism to handle contaminated data, i.e., real-world anomalies contaminate normal data during the training phase, as they all assume that the training data must be clean. We implement these baselines with their default model architecture, optimization procedure, and hyperparameters as suggested by the authors of the original papers. To ensure a fair comparison, we use the same datasets, experimental settings and evaluation protocols for TSAD-C and all compared methods.

\section{Implementation Details} \label{sec:implementation_details}

We set the masking ratio to match the anomaly ratio $\eta$ in $X_\text{train}$. Thus, for each dataset, $r$ is determined as the product of the observation length $L$ and $\eta$ as detailed in Table~\ref{tab:dataset}. We set $\Gamma$ to $L$ for each dataset, $d = 6$, $g=\frac{\Gamma}{d}$, $U = 128$, $\delta = 3$, $\zeta = 0.6$, $\xi_1 = 1$, $\xi_2 = 0.05$, and $\xi_3 = 0.5$. Additionally, we set $\beta_0 = 10^{-4}$, $\beta_T = 0.02$, $T = 50$, $\lambda_1 = 0.01$, and $\lambda_2 = 1.2$. The dimension of S4 states is fixed at 64. The batch size remains consistent at 4 for all datasets. The entire TSAD-C network is optimized via the AdamW optimizer \cite{loshchilov2017decoupled}, with a cosine learning rate scheduler \cite{loshchilov2016sgdr}, initialized with a learning rate of 8e-4. Model training would be early stopped if the validation loss fails to decrease for 20 consecutive epochs, with a maximum of 100 epochs. It is worth noting that to ensure a fair comparison, these hyperparameters are fixed across all experiments and datasets. All experiments are conducted on a single NVIDIA V-100 GPU (32 GB).

\section{Additional Visualizations} \label{sec:add_visualizations}

\begin{figure*}[t]
    \centering
    \begin{minipage}{0.48\textwidth}
        \centering
        \includegraphics[width=\linewidth]{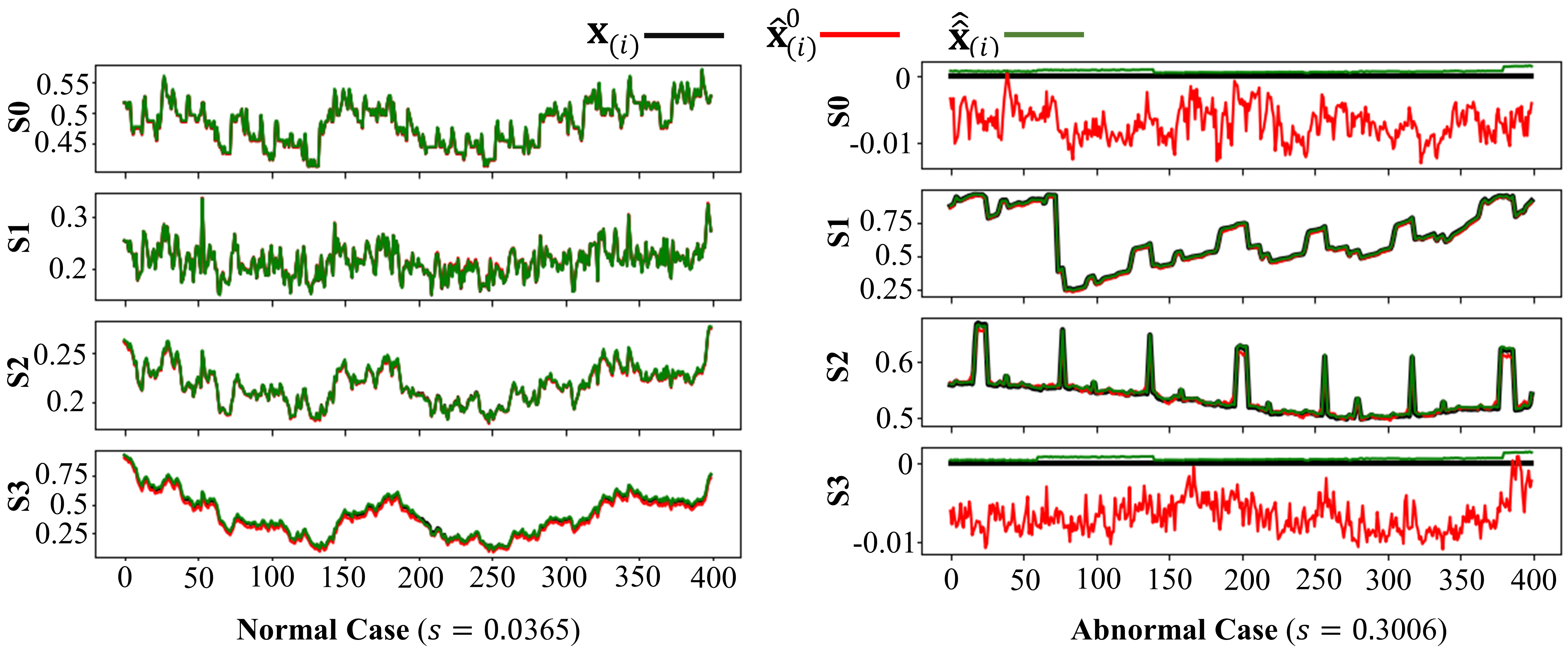}
        \small{(a) SMD}
    \end{minipage}
    \hfill
    \begin{minipage}{0.48\textwidth}
        \centering
        \includegraphics[width=\linewidth]{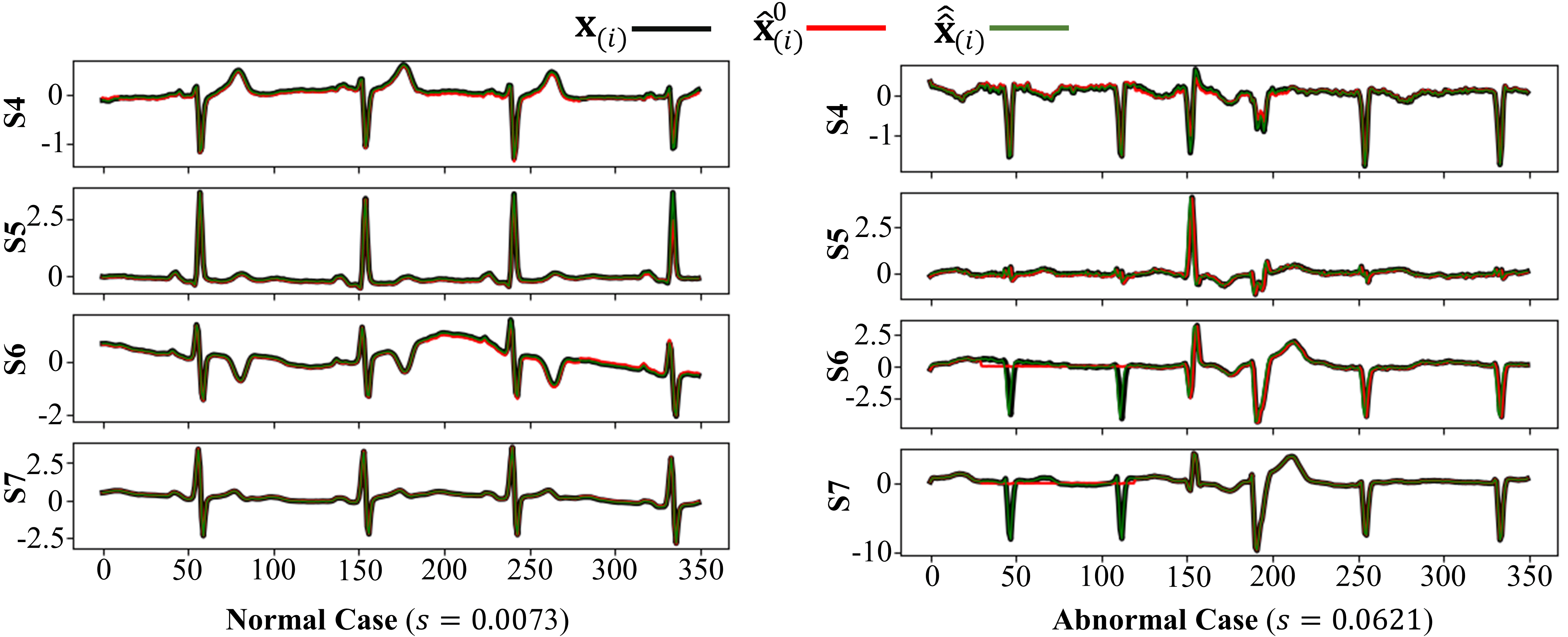}
        \small{(b) ICBEB}
    \end{minipage}
    \hfill
    \begin{minipage}{0.48\textwidth}
        \centering
        \includegraphics[width=\linewidth]{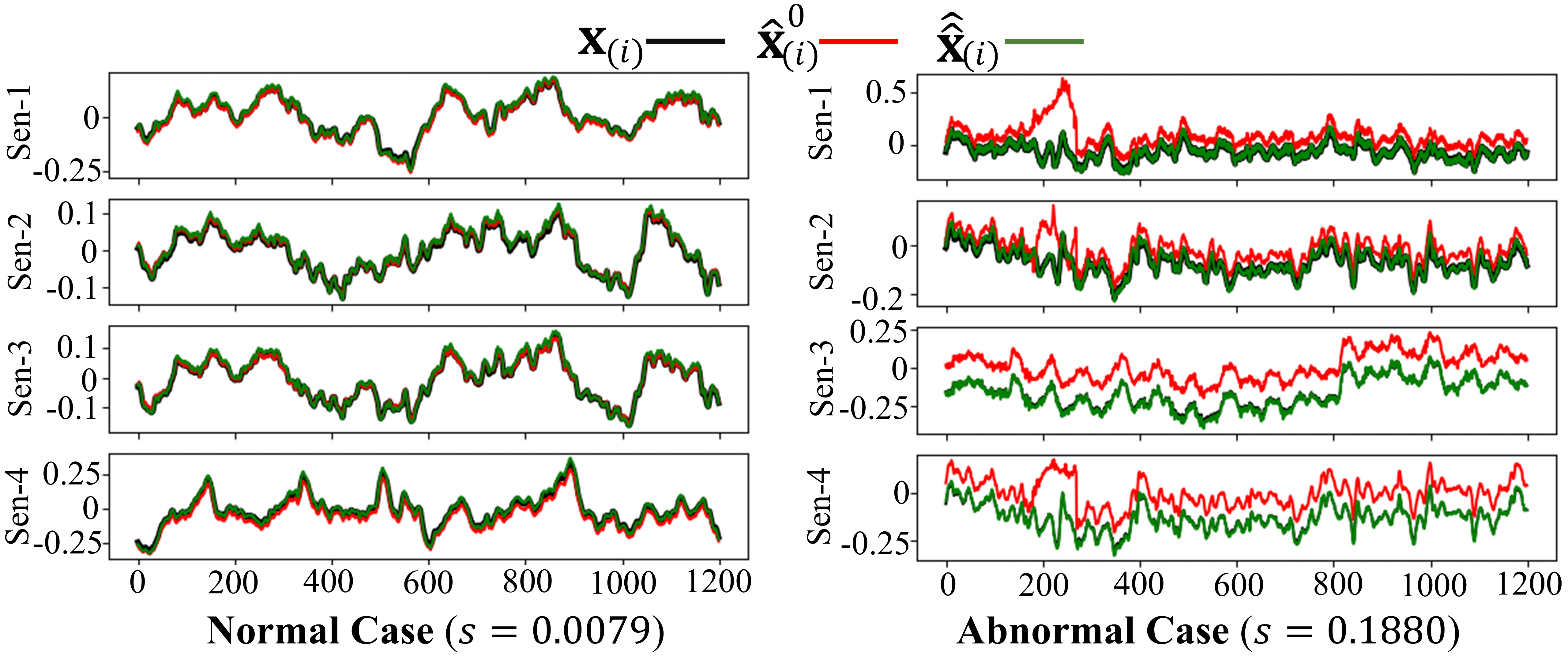}
        \small{(c) TUSZ}
    \end{minipage}
    \caption{Comparison between normal and abnormal cases for the masked segment in \br{(a) SMD and (b) ICBEB and (c) TUSZ.} The masked strategy used is BoM. Each case includes the ground truth $\mathbf{x}_{(i)}$, the decontaminated data $\hat{\mathbf{x}}_{(i)}^0$ and the reconstructed data $\mathbf{\doublehat{x}}_{(i)}$.}
    \label{fig:visualizations}
\end{figure*}

Figures \ref{fig:visualizations} (a), (b) and (c) show visualizations that compare the ground truth $\mathbf{x}_{(i)}$, decontaminated data $\hat{\mathbf{x}}_{(i)}^0$, and reconstructed data $\mathbf{\doublehat{x}}_{(i)}$ in SMD, ICBEB and TUSZ, respectively.
It is shown that $\hat{\mathbf{x}}_{(i)}^0$ and $\mathbf{\doublehat{x}}_{(i)}$ fit $\mathbf{x}_{(i)}$ very well in the normal case, leading to a lower $s$. Meanwhile, there are significant fluctuations of $\hat{\mathbf{x}}_{(i)}^0$ and $\mathbf{\doublehat{x}}_{(i)}$ compared to $\mathbf{x}_{(i)}$ in the abnormal case, resulting in a much higher $s$. These observations are consistent across all datasets. This shows the effectiveness of TSAD-C in distinguishing anomalies from normal data.

\section{Computational Cost and Scalability Analysis} \label{sec:scalability}

Due to the diversity in dataset characteristics and dimensions described in \textbf{Appendix \ref{sec:datasets}} and Table~\ref{tab:dataset}, in this section, we present dedicated discussions on TSAD-C's computational cost and scalability. Following the TSAD literature \cite{chen2022deep,li2021multivariate,wang2024fully}, we report the Floating-point Operations (FLOPs), the training time per epoch, the testing time per sample, and the detection performance of TSAD-C in Table~\ref{tab:computation}. Additionally, we conduct a detailed efficiency comparison between TSAD-C and comparable methods, i.e., InterFusion and IMDiffusion, which similarly handle both intra- and inter-variable dependencies, as discussed in \textbf{Appendix \ref{sec:baselines}}. It is shown that TSAD-C requires the fewest FLOPs. Notably, the training and testing times of TSAD-C are much faster than those of compared methods, while achieving significant improvements in detection performance. TSAD-C also maintains manageable training and testing durations, scaling reasonably as the dataset grows, when utilizing a single GPU. This efficiency is achieved despite significant dataset variations due to several reasons. First, our Decontaminator minimizes the noise error on only masked portions to have a simpler and more streamlined training process. The decontaminated data is then obtained in a \emph{single} step during the reverse process in the training phase, speeding up data preparation for subsequent modules -- a significant advantage for practical applications. Detailed studies are shown in Section~\ref{subsec:decontaminator_study}.

\renewcommand{\arraystretch}{1.4}
\setlength{\tabcolsep}{6pt}
\begin{table}[t]
    \caption{\br{Efficiency comparison between existing methods and TSAD-C in terms of FLOPs, the training time per epoch, the testing time per sample, and the detection performance (min: minutes, sec: seconds). The best scores are highlighted in bold.}}
    \centering
    \scalebox{0.8}{
    \begin{tabular}{c c c c c c c c}
        \hline
         & \multirow{2}{*}{\rotatebox[origin=c]{90}{\textbf{Method}}} & \multirow{2}{*}{\textbf{Dataset}} & \multirow{2}{*}{FLOPs} & \multirow{2}{*}{\makecell{Training time \\ per epoch \\ (min)}} & \multirow{2}{*}{\makecell{Testing time \\ per sample \\ (sec)}} & \multicolumn{2}{c}{Performance} \\ \cmidrule(lr){7-8} 
         & & & & & & F1 & APR \\ \hline 
         & \multirow{4}{*}{\rotatebox[origin=c]{90}{InterFusion}}  & SMD  & 4.6G & 4.2 &  35 & 0.383 & 0.490  \\ \cdashline{3-8} 
         & & ICBEB & 11.8G & 19.1 & 73 & 0.649 & 0.753  \\ \cdashline{3-8}
         & & DODH  & 20.4G & 29.4 &  91 & 0.418 &  0.559 \\ \cdashline{3-8}
         & & TUSZ  & 46.2G & 48.4 &  126 & 0.532  &  0.628 \\ \hline
         & \multirow{4}{*}{\rotatebox[origin=c]{90}{IMDiffusion}}  & SMD  & 5.8G & 5.4 & 69 & 0.426 &  0.533 \\ \cdashline{3-8} 
         & & ICBEB & 20.4G & 20.7 & 94 & 0.611 &  0.750 \\ \cdashline{3-8}
         & & DODH  & 32.2G & 31.5 & 116 & 0.544 &  0.651 \\ \cdashline{3-8}
         & & TUSZ  & 69.7G & 47.6 & 157 & 0.381 & 0.532 \\ \hline
         & \multirow{4}{*}{\rotatebox[origin=c]{90}{TSAD-C}}  & SMD  & 2.3G & 1.8 & 17 & \textbf{0.479} & \textbf{0.604} \\ \cdashline{3-8} 
         & & ICBEB  & 7.3G & 12.2 & 32 & \textbf{0.707} & \textbf{0.773} \\ \cdashline{3-8}
         & & DODH  & 12.2G & 19.3 & 66 & \textbf{0.652} & \textbf{0.728} \\ \cdashline{3-8}
         & & TUSZ  & 32.1G & 38.4 & 84 & \textbf{0.545} & \textbf{0.652} \\ \hline
         
    \end{tabular}}
    \label{tab:computation}
\end{table}

Additionally, as mentioned in Section~\ref{subsec:modeling}, since each observation can encompass thousands of timestamps (e.g., ICBEB with 6,000 timestamps, DODH with 7,500 timestamps and TUSZ with 12,000 timestamps), constructing a graph for every time step becomes inefficient and computationally demanding. To address this, we construct a graph over a defined time window, determined by the hyperparameter $g$, which aids in information aggregation. This strategy not only leads to a graph with reduced noise but also facilitates faster computations. Moreover, our graph module focuses on local connectivity patterns by $\delta$-nearest neighbors, which allow to scale efficiently because each node's computation is limited to its neighbors, reducing the overall computational complexity compared to fully-connected networks \cite{wang2024fully}. We also include the regularization terms in $\mathcal{L}_\text{graph}$ to regularize the graphs, as shown in Equation~\eqref{eq: loss_graph}. For instance, including $\mathcal{L}_\text{sparse}$ helps avoid overly connected graphs, which reduces computational costs for very large datasets like TUSZ. This regularization also aids in maintaining computational efficiency while preserving the graph’s ability to capture relevant data patterns. These aspects ensure TSAD-C's efficiency to varying dataset sizes without compromising on detection performance.

\end{document}